\documentclass[review,12p]{elsarticle}

\usepackage{sidecap}
\usepackage{lineno,hyperref}
\usepackage[T1]{fontenc}
\usepackage{graphicx}
\usepackage{setspace}	
\usepackage{subfloat}
\usepackage{float}
\usepackage{stfloats}
\usepackage[linesnumbered,ruled,vlined]{algorithm2e}
\usepackage[toc,page]{appendix}
\usepackage{adjustbox}
\usepackage{multirow}
\usepackage{hhline}
\usepackage{gensymb}
\usepackage{algorithmic,algorithm2e,float}
\usepackage{times}
\usepackage{epsfig}
\usepackage{amssymb}
\usepackage{amsmath,amsfonts,bm}
\usepackage{caption}
\usepackage[linesnumbered,ruled,vlined]{algorithm2e}
\usepackage{pifont}
\usepackage{fixmath}
\usepackage{epstopdf}
\usepackage{cleveref}
\usepackage{makecell}
\usepackage{subfigure}   
\usepackage[table]{xcolor}
\usepackage{lipsum}

\makeatletter
\def\ps@pprintTitle{%
   \let\@oddhead\@empty
   \let\@evenhead\@empty
   \def\@oddfoot{\reset@font\hfil\thepage\hfil}
   \let\@evenfoot\@oddfoot
}
\makeatother

\begin{document}

\begin{frontmatter}

\newcommand{\cmark}{\ding{51}}%
\newcommand{\xmark}{\ding{55}}%
\newcommand{\stimes}{{\times}}
\newcommand{\specialcell}[2][c]{%
\begin{tabular}[#1]{@{}c@{}}#2\end{tabular}}

\title{A Paired Sparse Representation Model for Robust Face Recognition from a Single Sample}
%\title{Domain-invariant Face Recognition from a Single sample using an Integrated Sparse Representation Model}
%\title{Robust Face Recognition from a Single Still using a Synthetic plus Variation Model}

\author{Fania Mokhayeri\corref{cor1}}
\ead{fmokhayeri@livia.etsmtl.ca}
\cortext[cor1]{Corresponding author}

\author{Eric Granger\corref{cor2}}
\ead{eric.granger@etsmtl.ca}

\address{Laboratoire d'imagerie, de vision et d'intelligence artificielle (LIVIA) \\ \'Ecole de Technologie Sup\'erieure, Universit\'e du Qu\'ebec, Montreal, Canada}

%**********************************
\begin{spacing}{1.2}
\begin{abstract}
Sparse representation-based classification (SRC) has been shown to achieve a high level of accuracy in face recognition (FR). However, matching faces captured in unconstrained video against a gallery with a single reference facial still per individual typically yields low accuracy. For improved robustness to intra-class variations, SRC techniques for FR have recently been extended to incorporate variational information from an external generic set into an auxiliary dictionary. Despite their success in handling linear variations, non-linear variations (e.g., pose and expressions) between probe and reference facial images cannot be accurately reconstructed with a linear combination of images in the gallery and auxiliary dictionaries because they do not share the same type of variations.
In order to account for non-linear variations due to pose, a paired sparse representation model is introduced allowing for joint use of variational information and synthetic face images. The proposed model, called \textit{synthetic plus variational model}, reconstructs a probe image by jointly using (1) a variational dictionary and (2) a gallery dictionary augmented with a set of synthetic images generated over a wide diversity of pose angles. The augmented gallery dictionary is then encouraged to pair the same sparsity pattern with the variational dictionary for similar pose angles by solving a newly formulated simultaneous sparsity-based optimization problem.   
Experimental results obtained on Chokepoint and COX-S2V datasets, using different face representations, indicate that the proposed approach can outperform state-of-the-art SRC-based methods for still-to-video FR with a single sample per person. 
%it allows to achieve a high level of accuracy without having to collect a large number of reference training samples.
\end{abstract}

%*************************************************************

\end{spacing}

\end{frontmatter}

%*************************************************************
\section{Introduction}
\label{intro}
%*************************************************************
Video-based face recognition (FR) has attracted a considerable amount of interest from both academia and industry due to the wide range applications as found in surveillance and security. In contrast to FR systems based on still images, an abundance of spatio-temporal information can be extracted from target domain videos to contribute in the design of discriminant still-to-video FR systems.

Sparse Representation-based Classification (SRC) techniques can provide an accurate and cost-effective solution in many video FR applications when there are a sufficient number of reference training images per each person under controlled condition~\cite{Wright1, xu, xu2017}. 
However, single sample per person (SSPP) problems are common in video-based security and surveillance applications, as found in, e.g., biometric authentication and watch-list screening~\cite{farshad, Dewan}. For example, still-to-video FR systems are typically designed using only one reference still image per individual in the source domain, and then faces captured with video surveillance cameras in target domain are matched against these reference stills~\cite{S2, S3}. Additionally, when faces are captured under challenging uncontrolled conditions, they may vary considerably according to pose, illumination, occlusion, blur, scale, resolution, expression, etc. In such cases, using SRC techniques often associated with limited robustness to intra-class variations, and a lower recognition rate. 

State-of-the-art approaches designed to address SSPP problems in SRC-based FR systems can be roughly divided into three categories: (1) image patching methods, where the images are partitioned into several patches~\cite{Zhu, gaor}, (2) face synthesis technique to expand the gallery dictionary~\cite{mokhayeri, hu}, and (3) generic learning methods, where a genetic training set\footnote{A generic set is defined as an auxiliary set comprised of many facial video ROIs from unknown individuals captured in the target domain.} is used to leverage variational information from an auxiliary generic set of images to represent the differences between probe and gallery images ~\cite{Wei, deng2018}. Indeed, similar intra-class variations may be shared by different individuals in the generic set and reference regions of interest (ROIs) in the gallery. Moreover, a generic set can be easily collected during operations or some camera calibration process, and encode subtle knowledge on faces appearing in the operational environment. One of the pioneering techniques in generic learning is extended SRC (ESRC)~\cite{Deng}, which manually constructs an auxiliary variational dictionary from a generic set to accurately represent a probe face with unknown variations from the target domain. ESRC was subsequently generalized to employ different sparsity for identity and variational parts in sparse coefficients~\cite{Li}, and to learn the variational dictionary that accounts for the relationship between the reference gallery and external generic set~\cite{Yang}.

% Challenges 
Although leveraging intra-class variations from a generic set has been shown to improve robustness to some linear facial variations, it cannot accurately address non-linear facial variations (e.g., pose and expression) between reference still ROIs in the source domain and probe videos ROIs captured in real-world capture conditions in the target domain. Indeed, non-linear variations are not additive nor sharable. For instance, a probe video ROI with various lighting can be recovered with a linear combination of an image with a natural lighting and its corresponding illumination component. However, a probe ROI with a profile view cannot be accurately reconstructed with a linear combination of frontal view ROIs in gallery dictionary and profile view ROIs in the auxiliary dictionary because they do not share the same type of variations. Non-linear facial variations between still and video ROIs make it difficult to represent a probe image using a linear combination of reference and generic set images. Another concern with ESRC is the large manually designed auxiliary dictionary (obtained via random selection in the generic set) which is computationally expensive.
To address these concerns, we focus on two issues: (1) how to represent a probe image under non-linear variations with a linear combination of reference set and generic set, (2) how to  design a discriminative dictionary, and (3) how to yield a robust representation with a minimum number of images.

\begin{figure}[!tp]
	\centering 
	\advance\leftskip 0.1cm
	\includegraphics[width=.9\textwidth]{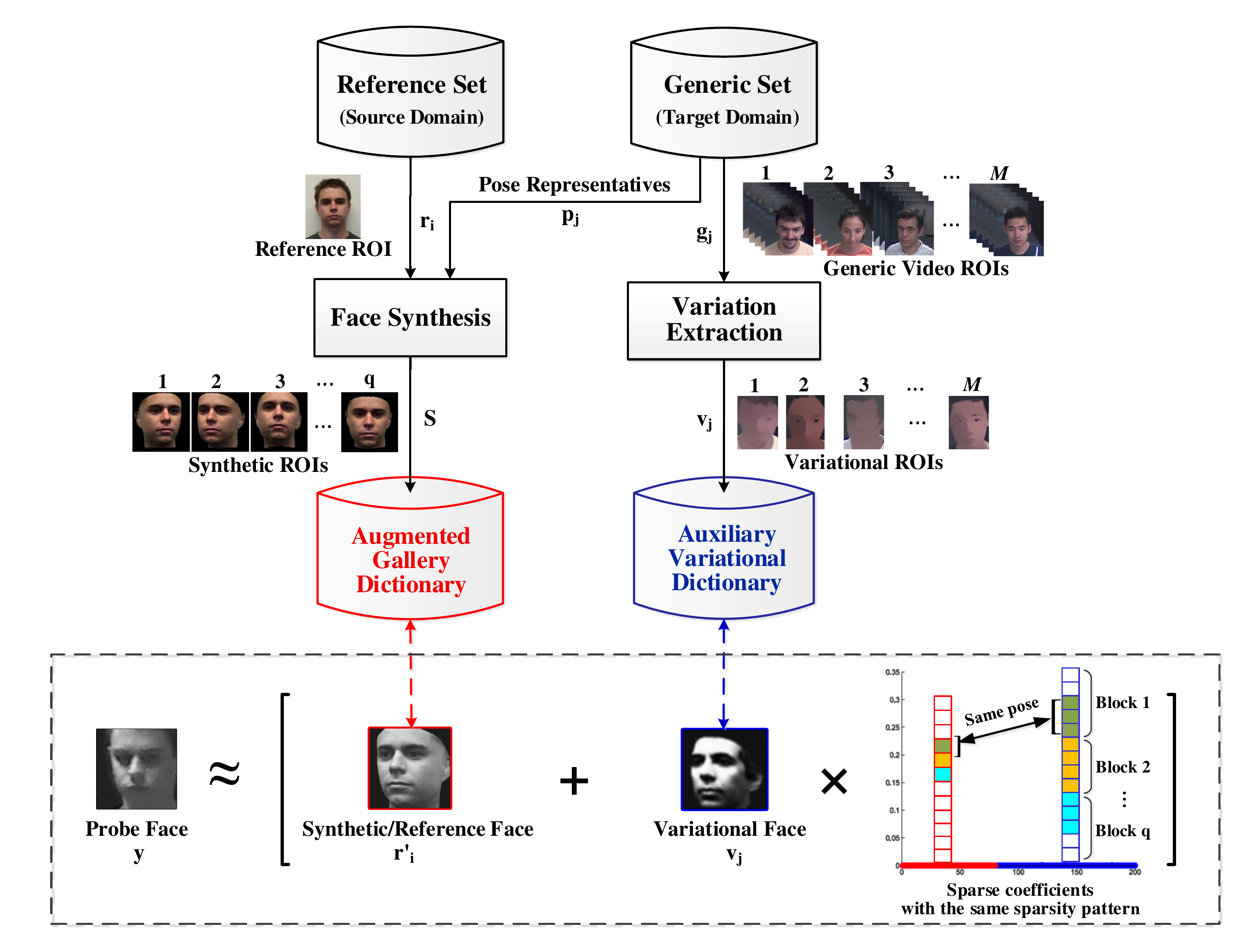}
	\caption{\small Overall architecture of the proposed approach. The gallery dictionary is augmented with a diverse set of synthetic images and the auxiliary variational dictionary co-jointly encode non-linear variations in appearance. Sparse coefficients within each dictionary share the same sparsity pattern in terms of pose angle.} \label{fig1}
\end{figure}

% Proposal
In this paper, a paired sparse representation framework referred as the \textit{synthetic plus variational model} (S+V) is proposed to address the problem of non-linear pose variations by increasing the range of pose variations in the gallery dictionary. Since collecting a large database with a wide variety of views is extremely expensive and time-consuming, a set of synthetic face images under representative pose are generated. As illustrated in Fig.~\ref{fig1}, a probe video ROI is reconstructed using an auxiliary dictionary as well as a gallery dictionary augmented with a set of synthetic face images generated under a representative diversity of azimuth angles. The proposed sparse model not only allows probe image to be represented by the atoms of both augmented and auxiliary dictionaries, but also restricts the selected atoms to be combined with the same viewpoint, thus providing an improved representation. 

Under this model, facial ROIs from trajectories in the generic set are clustered in the captured condition space (defined by pose angle) by applying row sparsity~\cite{Elhamifar2}. The auxiliary variational dictionary with block structure is designed using intra-class variations as subsets the pose clusters. Following this, the gallery dictionary is augmented with the synthetic face images generated from the original reference image in the source domain, where the rendering parameters are estimated based on the center of each cluster in the target domain. By introducing a joint sparsity structure, the pose-guided augmented gallery dictionary is encouraged to share the same sparsity pattern with the auxiliary dictionary for the same pose angles. Each synthetic facial ROI in the augmented gallery dictionary is thereby combined with approximately the same facial viewpoint in the variational dictionary in a joint manner \cite{rakotomamonjy}. During the operation, each input probe face captured in videos is represented by a linear combination of ROIs from a same person and same pose in the augmented gallery dictionary as well as the intra-class variations from a same pose in the auxiliary variational dictionary. In this framework, the auxiliary dictionary models the linear variations (such as illumination changes, different occlusion levels) and non-linear pose variation are modeled by augmented gallery dictionary. Note that the S+V model is paired across different domains in the enrollment stage. The main contributions of this paper are:
\begin{itemize} 
\item A generalized sparse representation model for still-to-video FR, using generic learning and data augmentation to represent both linear and non-linear variations based on only one reference still ROI;
\item A simultaneous optimization technique to encourage pairing between each synthetic profile image in the augmented gallery dictionary and a similar view in the auxiliary dictionary; 
\item An efficient SRC method to design a compact augmented dictionary using row sparsity. 
\end{itemize}
This paper extends our preliminary investigation of synthetic plus variational models  \cite{Fania_FG2019} in several ways, in particular with: (1) a comprehensive analysis of dictionary design and of selection of representative face exemplars; (2) a detailed description of the proposed joint sparsity structure; and (3) more experimental results and interpretations, including results with deep facial representations, an ablation study and complexity analysis.

For proof-of-concept validation, a particular implementation of the proposed SRC technique for still-to-video FR is considered where representative pose angles are selected by applying clustering on the generic set. The original and synthetic ROIs rendered under these pose angles are employed to design an augmented gallery dictionary, while the pose clusters of video ROIs are exploited to design an auxiliary variational dictionary with block structure. The simultaneous sparsity constraint is then applied to both dictionaries to improve the discrimination power of the dictionaries. Moreover, since most state-of-the-art FR methods rely on Convolution Neural Network (CNN) architectures such as ResNet~\cite{He} and VGGNet~\cite{simonyan}, the model is fed with CNN features extracted from the atoms of dictionaries \cite{Gao, Cai}, in order to further improve still-to-video FR accuracy. Performance of the SRC implementation is evaluated on two public video FR databases -- Chokepoint~\cite{Wong} and COX-S2V~\cite{Huang3}. 

The rest of the paper is organized as follows. Section~\ref{RW} provides a brief review for SRC methods that employ generic learning to address SSPP problems. Section~\ref{S+V} describes the proposed S+V model. Section~\ref{FR} presents a particular implementation of the S+V model for still-to-video FR system. Finally, Sections~\ref{experiment} and \ref{results} describe the methodology and experimental results, respectively.

%*************************************************************
\section{Background on Sparse Modelling for Still-to-Video FR} 
\label{RW}
%*************************************************************

%\subsection{Notation:}
In the following, the set $\mathbf{D} =  \{ \mathbf{r}_1,\mathbf{r}_2, \dotsc ,\mathbf{r}_k \} \in \mathbb{R}^{d \times k}$ composed of $1$ reference still ROI belonging to one of $k$ different classes, $d$ is the number of pixels or features representing a ROI and $n$ is the total number of reference still ROIs. 
The set $\mathbf{G} = \{\mathbf{g}_1, \mathbf{g}_2 \dotsc ,\mathbf{g}_m \} \in \mathbb{R}^{d \times m}$ denotes the auxiliary generic set composed of $m$ external generic images of unknown persons captured in the target domain. The set $\mathbf{V} = \{\mathbf{v}_1, \mathbf{v}_2,  \dotsc ,\mathbf{v}_m \} \in \mathbb{R}^{d \times m}$ denotes the auxiliary variational dictionary composed of $m$ intra-class variations extracted from $\mathbf{G}\in \mathbb{R}^{d \times m}$.
 
%****************************************
\subsection{Sparse Representation-based Classification (SRC):} 
% Given an input data matrix, sparse coding aims to find a set of basis vectors (i.e., dictionary) that capture high-level semantics, and the sparse coordinates with respect to the dictionary.
Given a probe image $\mathbf{y}$, SRC represents $\mathbf{y}$ as a sparse linear combination of a reference set $\mathbf{D} \in \mathbb{R}^{d \times k}$. SRC uses the $\ell_1$-minimization to regularize the representation coefficients. More precisely, SRC derives the sparse coefficient $\pmb{\alpha}$ of $\mathbf{y}$ by solving the following $\ell_1$-minimization problem: 
\begin{equation}
\min_\alpha \| \mathbf{y} - \mathbf{D}\pmb{\alpha} \|_2^2  +  \lambda \|\pmb{\alpha}\|_1.
\label{eq1}
\end{equation} 
where $\lambda$ is a regularization parameter, and $\lambda > 0$. After the sparse vector of coefficients $\pmb{\alpha}$ is obtained, the probe image $\mathbf{y}$ is recognized as belonging to class $k^*$ if it satisfies:  
\begin{equation}
k^*  = \underset{k}{\arg\min}  \| \mathbf{y - D} {\gamma}_{k} (\pmb{\alpha}) \|_2.    
\label{eq2}
\end{equation}
where ${\gamma_{k}}$ is a vector whose only nonzero entries are the entries in $\pmb{\alpha}$ that are associated with class $k$. 
%That is, the probe image $\mathbf{y}$ will be assigned to the class with the minimum class-wise reconstruction error.
SRC is based on the idea that a probe image $\mathbf{y}$ can be best linearly reconstructed by the columns of $\mathbf{D}_{k^*}$ if it belongs to class $k^*$. As a result, most non-zero elements of $\pmb{\alpha}$ will be associated with class $k^*$, and  $\| \mathbf{y - D} {\gamma}_{k^*} (\pmb{\alpha}) \|_2$ yields the minimum reconstruction error. An important assumption of SRC is that it requires a large amount of reference training images to form an over-complete dictionary. However, in many practical applications, the number of labeled reference images are limited, and SRC accuracy declines in such cases~\cite{Wright1}.

%****************************************
\subsection{SRC through Generic Learning:} 
Since the facial variations share much similarity across different individuals, an external generic set with multiple images of unknown persons as they appear in the target domain can provide discriminant information on intra-class variations. These additional variations can enrich the gallery diversity, especially in SSPP scenarios. The general model solves the following minimization problem:
\begin{equation}
\min_{\alpha,\beta} \bigg\|\mathbf{y} -\mathbf{[D, V]
	\begin{bmatrix}
	\pmb{\alpha} \\ \pmb{\beta}
	\end{bmatrix} 
}\bigg\|_a^a  +  \lambda 
\bigg\|
\begin{bmatrix}
\pmb{\alpha} \\ \pmb{\beta}
\end{bmatrix} \bigg\|_b^b .
\label{eq3}
\end{equation}
where $\pmb{\alpha}$ is a sparse vector that selects a limited number of variant bases from the gallery dictionary $\mathbf{D}$, and $\pmb{\beta}$ is another sparse vector that selects a variant bases from the auxiliary variational dictionary $\mathbf{V}$, $a \in \{ 1,2 \}$, $b \in \{ 1,2 \}$ and $\lambda > 0$.
The variant bases can be estimated by subtracting the natural (original) image of a class from other images of the same class, the difference from the class centroid, and pairwise difference. The probe image $\mathbf{y}$ is recognized as belonging to class $k^*$ if it satisfies:  
\begin{equation}
k^* =  \underset{k}{\arg\min} \bigg\|\mathbf{y} -\mathbf{[D, V]
	\begin{bmatrix}
	{\gamma}_k (\pmb{\alpha}) \\ \pmb{\beta}
	\end{bmatrix}  
}\bigg\|_a^a .
\label{eq4}
\end{equation}
where ${\gamma}_k$ is reused as a matrix operator. % The most non-zero elements of $\alpha$ will be mainly presented in the non-zero elements of ${\gamma}_{k}(\alpha)$ and thus result in the minimum reconstruction error.

Deng \textit{et al}.~\cite{Deng} introduced extended SRC (ESRC), which manually designs an auxiliary dictionary (through random selection from a generic set) to accurately represent a probe face with unknown variations from the target domain. The model of Eq.~\ref{eq4} degenerates to the ESRC model when $a = 2$ and $b = 1$. Motivated by ESRC, Yang \textit{et al}.~\cite{Yang} proposed the sparse variation dictionary learning (SVDL) model to learn the variational dictionary by accounting for the relationship between the reference gallery and external generic set. 
A robust auxiliary dictionary learning (RADL) technique was proposed in~\cite{Wei} that extracts representative information from external data via dictionary learning without assuming the prior knowledge of occlusion in probe images. In~\cite{farshad}, variational information from the target domain was integrated with the reference gallery set through domain adaptation to enhance the facial models for still-to-video FR. A new approach is proposed to learn a kernel SRC model based on a virtual dictionary and the original training set \cite{fan2018}.
Authors in \cite{deng2018} developed a superposed linear representation classifier to cast the recognition problem by representing the test image in term of a superposition of the class centroids and the shared intra-class differences. A local generic representation-based (LGR) framework for FR with SSPP was proposed in~\cite{Zhu}. It builds a gallery dictionary by extracting the patches from the gallery database, while an intra-class variation dictionary is formed by using an external generic set to predict the possible facial variations (\textit{e.g.}, illuminations, pose, and expressions). In order to address non-linearity, authors in~\cite{fan} used a nonlinear mapping to transform the original reference data into a high dimensional feature space, which is achieved using a kernel-based method. A customized SRC (CSR) had been proposed to leverage the different sparsity of identity and variational parts in sparse coefficients, and to assign different parameters to their regularization terms~\cite{Li}. In~\cite{yang2017}, a joint and collaborative sparse representation framework was presented that exploits the distinctiveness and commonality of different local regions. 
A novel discriminative approach is proposed in \cite{lin2018}, in which a robust dictionary is learned from diversities in training samples, generated by extracting and generating facial variations. In \cite{xie2019} feature sparseness-based regularization is proposed to learns deep features with better generalization capabilities. In this paper, the regularization is integrated into the original loss function, and optimized with a deep metric learning framework. Authors in \cite{luo2019} propose a novel multi-resolution dictionary learning method for FR that provides multiple dictionaries -- each one associated with a resolution -- while encoding the similarity of representations obtained using different dictionaries in the training phase.
3D Morphable Model (3DMM), proposed by Blanz and Vetter \cite{blanz1}, has been widely used to synthesize new face images from a single 2D face image.
The 3DMM is expanded by adopting a shared covariance structure to mitigate small sample estimation problems associated with data in high dimensional spaces \cite{koppen2018}. It models the global population as a mixture of Gaussian sub-populations, each with its own mean value. Finally, an efficient deep learning model for face synthesis is proposed in \cite{jiao2018} which is does no rely on complex optimization.

The aforementioned techniques work well in video-based FR. However, they neglect the impact of non-linear variations between probe images and facial images in the gallery and auxiliary dictionaries. To account for the non-linearities, particularly  pose variations, the range of viewpoints represented in the gallery dictionary should be increased to represent the probe image with the same view gallery and variations, and thereby compensate the non-linear pose variations. Additionally, the sparsity pattern should ensure the correlation between the gallery and variational dictionaries in terms of pose angles.

%*************************************************************
\section{The Proposed Approach - A Synthetic plus Variational Model} 
\label{S+V}
%*************************************************************

In this section, a new sparse representation model -- called the \textit{Synthetic plus Variational} (S+V) model -- is proposed to overcome issues related to the non-linear pose variations with conventional and ESRC model.
% Sparsity-based FR techniques that employ generic intra-class variations to overcome SSPP has shown promising results; however, they neglected the impact of nonlinear pose variations between the probe sample and training set in representing the face intra-class variation information. 
SRC techniques commonly assumed that frontal and profile views share the same type of variations. To address this limitation, we increase the range of pose variations of gallery dictionary to represent the probe with the same view gallery and variations, and accordingly compensate the non-linear pose variations.

\begin{figure}
    \centering 
    \advance\leftskip 0.1cm
	\includegraphics[width=.97\textwidth]{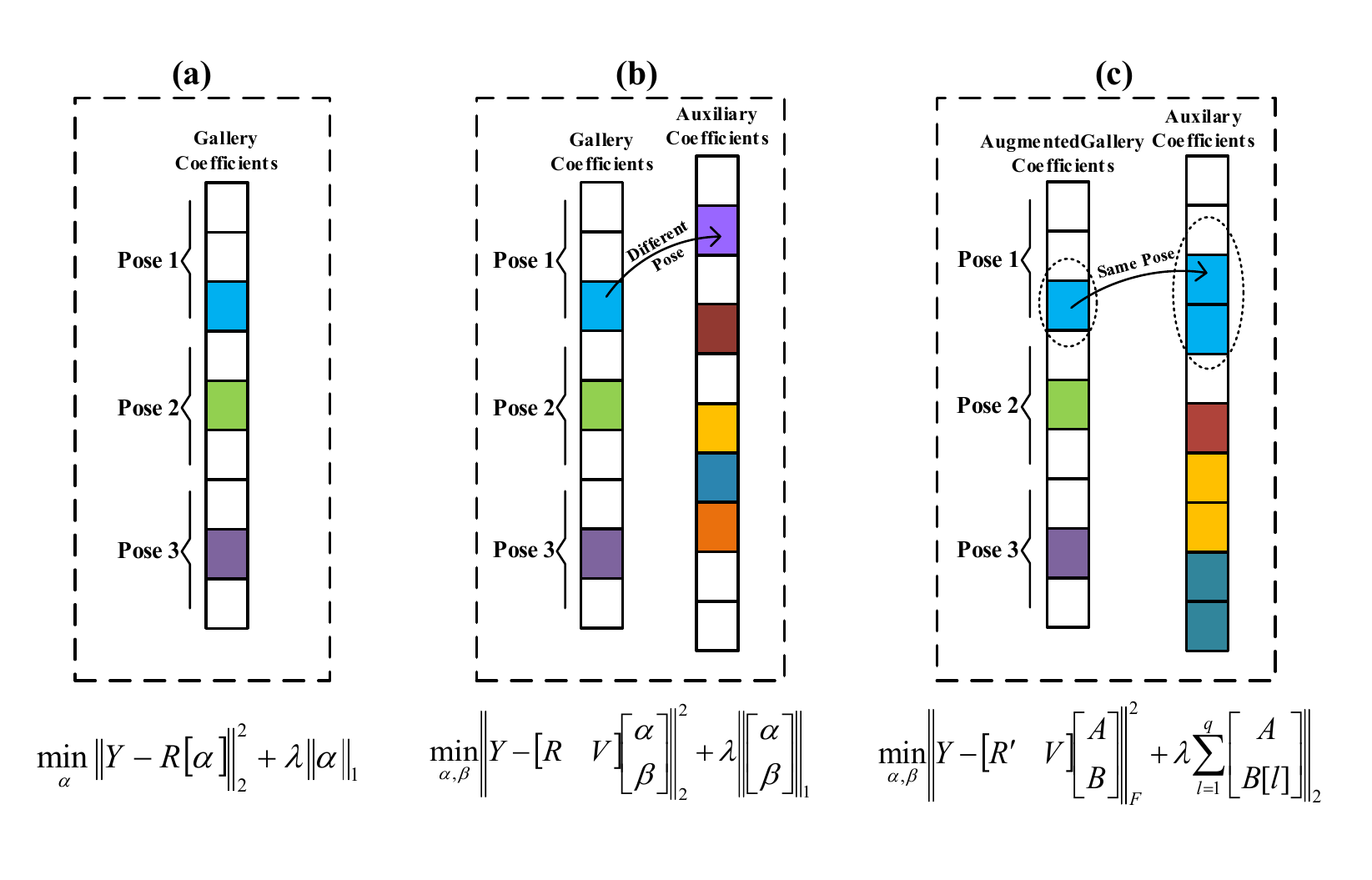}
    \vspace{-.2cm}
	\caption{\small A comparison of the coefficient matrices for three sparsity models: (a) Independent sparsity (SRC) with a single dictionary, (b) Extended sparsity (ESRC) with two dictionaries, and (c) Paired extended sparsity (S+V model) with pair-wise correlation between two dictionaries where the sparse coefficients of same poses share the same sparsity pattern. Each column represents a sparse coefficient vector and each square block denotes a coefficient value. 
	%White blocks denote zero values, whereas color blocks stand for nonzero values. 
	%The column dimension represents the number of poses in the gallery grouped by an individual.
	} 
    \label{fig2}
\end{figure}

The proposed S+V model exploits two dictionaries including (1) an augmented gallery dictionary containing the original reference still ROI of each individual as well as their synthetic profile ROIs (with diverse poses) enrolled to the still-to-video FR system, and (2) an auxiliary variational dictionary which contains variations from the target domain that can be shared by different persons. Two dictionaries are correlated by imposing the simultaneous sparsity prior that force the augmented gallery dictionary to pair the same sparsity pattern with the auxiliary dictionary for the same pose angles. In this manner, each synthetic profile image in the augmented gallery dictionary is combined with approximately the similar view in the auxiliary dictionary. Fig.~\ref{fig2} gives an illustrative example that compares the sparsity structure of SRC, ESRC and S+V model. The rest of this section presents more details on the dictionary design and encoding process with the S+V model.

%**********************************************
\subsection{Dictionary Design:}\label{CCC}
%**********************************************
In order to design the gallery and auxiliary dictionaries, the representative pose angles are determined by characterizing the capture conditions from a large generic set of video ROIs in the pose space (estimations of pitch, roll, and yaw). Prior to operation, e.g., during a camera calibration process, facial ROIs are isolated in facial trajectories from the videos of unknown persons captured in the target domain. A representative set of video ROIs are selected by applying row sparsity regularized optimization program on facial trajectories in the captured condition space defined by pose angles. Next, the variational information of the generic set with multi-samples per person are extracted to form an auxiliary dictionary based on the subsets of the pose clusters. A compact set of synthetic images is then generated from the reference set in the source domain based on the information obtained from the center of each cluster in the target domain, called pose representatives, and integrated into the gallery dictionary to  enrich  the  diversity  of  the  gallery  set. Two dictionaries are correlated by imposing the simultaneous sparsity prior that force the same sparsity patterns among the multiple sparse representation vectors in the augmented and auxiliary dictionaries in terms of pose angles. Finding representative poses not only are employed to make a pair-wise correlation between the dictionaries but also can save time and memory and improve the recognition performance due to preventing over-fitting.
Inspired by~\cite{Elhamifar2, Elhamifar3}, the representative selection problem is formulated as a row sparsity regularized trace minimization problem where the objective is to find a few representatives (exemplars) that efficiently represent the collection of data points according to their dissimilarities. 

The proposed model allows to select pose representatives from a collection of ${N}$ pose samples. The pose angles are estimated using the discriminative response map fitting method~\cite{zhx} which is a state-of-the-art method for accurate fitting, suitable for handling occlusions and changing illumination conditions. The estimated head pose for the $j^\text{th}$ video ROI ($\mathbf{g}_j$) in the generic set is defined as $\pmb{\theta}_j = (\pmb{\theta}^{pitch}_j, \pmb{\theta}^{yaw}_j, {\pmb\theta}^{roll}_j)$. Euler angles $\pmb{\theta}^{pitch}$, $\pmb{\theta}^{yaw}$, and $\pmb{\theta}^{roll}$ are used to represent roll, yaw and pitch rotation around $X$ axis, $Y$ axis, and $Z$ axis of the global coordinate system, respectively. 
The set of dissimilarities $\{d_{ij}: i,j=1,..., k \}$ between every pair of pose data points are then calculated by using the Euclidean distance, which indicates how well the data point $i$ is suited to be an exemplar of data point $j$. The dissimilarities are arranged into matrix:
\begin{equation}
\mathbf{D} \triangleq \left[ \begin{matrix} \mathbf{d}_1^T \\ \vdots \\ \mathbf{d}_N^T \end{matrix}\right] 
=
\left[ \begin{matrix}
d_{11} & d_{12} & \cdots & d_{1k}\\
\vdots & \vdots & \ddots & \vdots\\
d_{k1} & d_{k2} & \cdots & d_{kk}\\
\end{matrix}\right]  \in \mathbb{R}^{k \times k} ,
\label{eq5}
\end{equation}
\noindent where $\mathbf{d}_{i}$ denotes the $i^{\textit{th}}$ row of $\mathbf{D}$. 
Variables $z_{ij}$ are associated with dissimilarities $d_{ij}$, and organized into matrix of the same size as:
\begin{equation}
\mathbf{Z} \triangleq \left[ \begin{matrix} \mathbf{z}_1^T \\ \vdots \\ \mathbf{z}_N^T \end{matrix}\right] 
=
\left[ \begin{matrix}
z_{11} & z_{12} & \cdots & z_{1k}\\
\vdots & \vdots & \ddots & \vdots\\
z_{N1} & z_{N2} & \cdots & z_{kk}\\
\end{matrix}\right]  \in \mathbb{R}^{k \times k} ,
\label{eq6}
\end{equation}\\
where $z_i \in \mathbb{R}^k$ denotes the $i^{th}$ row of $\mathbf{z}$. $z_{ij}$ is the probability that data point $i$ is representative for data point $j$, and $z_{ij} \in [0, 1]$. The row sparsity regularized trace minimization algorithm is applied on matrix $\mathbf{Z}$ to select some representative exemplars that can suitably encode pose data according to dissimilarities as follows:
\begin{equation}
\min \sum_{j=1}^k \sum_{i=1}^k d_{ij} z_{ij} +  \eta \sum_{i=1}^k \big\| z_i \big\|_q,
\label{eq7}
\end{equation}
subject to:
\begin{equation*}
\quad z_{ij} \geq {0},  \quad \forall i,j; \quad \sum_{i=1}^k z_{ij}=1, \quad \forall j,
\label{eq71}
\end{equation*}
where the parameter $\eta > 0$ sets the trade-off between these two terms.
%As we change the regularization parameter $\eta $ in \ref{eq7}, the number of representatives found by the algorithm changes. For small values of $\eta, where we put more emphasis on better encoding data points via representatives, we obtain more representatives. On the other hand, for large values of $\eta, where we put more emphasis on the row sparsity of $\mathbf{Z}$, we select a small number of representatives.

Once this optimization problem (Eq.~\ref{eq7}) has been solved, one can find the representative indices from the nonzero rows of $\mathbf{Z}$. The clustering of data points into $K$ clusters, associated with $K$ representatives, is obtained by assigning each data point to its closest representative. In particular, if \{ $i_1; \dotsc ; i_q$ \} denote the indices of the representatives, data point $j$ is assigned to the pose representative $\theta(j)$ such that $\theta(j) = \arg \min_{\ell \in { \{i_1; \dotsc ; i_q\} }}d_{\ell j}$.
% the solution Z gives the probability that each data point associated with each one of the representatives, which also provides a soft clustering of data points to the representatives.

The auxiliary dictionary is designed based on these pose clusters, where each cluster forms a block in the dictionary. The pose angle of representative video ROI of each pose cluster, referred as pose exemplar, is used as rendering parameter to generate synthetic face images with varying poses using off-the-shelf 3D face models~\cite{blanz1,Tran1,Tran2}. In this way, $q$ synthetic profile faces, $\mathbf{S} = \{ \mathbf{S}_{i} : i = 1, \dotsc, k \}$,  are generated under the representative pose angles from a given single still face image where $\mathbf{S}{i} = \{ \mathbf{s}_1^{i}, \mathbf{s}_2^{i}, \dotsc, \mathbf{s}_q^{i} \} \in \mathbb{R}^{d \times q}$.

The augmented gallery dictionary  $\mathbf{D}^\prime = \{ \mathbf{D}_{i}^\prime:i= 1,\dotsc, k \}$, is formed by merging each still ROI of reference set with $q$ synthetic images rendered w.r.t. representative pose exemplars, where here $\mathbf{D}_i^\prime = \{\mathbf{r}_1, \mathbf{s}_1^{i}, \mathbf{s}_2^{i}, \dotsc, \mathbf{s}_q^{i} \} \in \mathbb{R}^{d \times (1+q)}$.

%*************************************************************
\subsection{Synthetic Plus Variational Encoding:}
\begin{figure}[!t]
 	\centering 
    \vspace{-.2cm}
	\advance\leftskip -1.1cm
 	\includegraphics[width=1.2\textwidth]{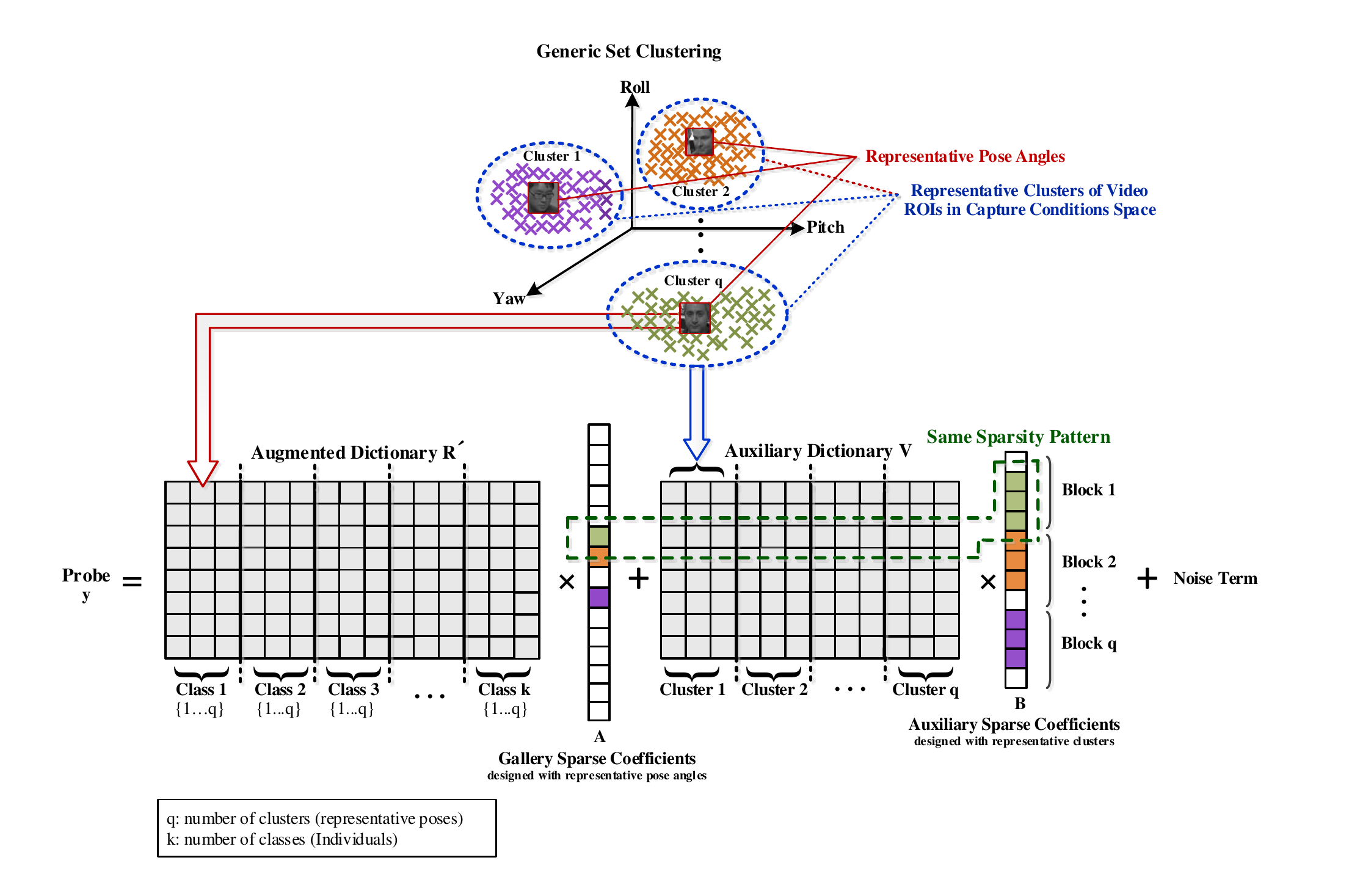}
 	\vspace{-.5cm}
    \caption{\small An illustration of sparsity pattern with the S+V model based on clustering results in the pose space. Each column represents a sparse representation vector, each square denotes a coefficient and each matrix is a dictionary.}
 	\label{fig3}
 \end{figure}

%************************************************************
With the S+V model (see Fig.~\ref{fig3}), each probe video ROI is seen as a combination of two different sub-signals in the augmented gallery dictionary and auxiliary variation dictionary in the linear additive model:
\begin{equation}
\mathbf{y} = \mathbf{D^\prime}\pmb{\alpha} + \mathbf{V}\pmb{\beta}  + e  ,       
\label{eq5}
\end{equation}
where $\mathbf{D^\prime} \in \mathbb{R}^{d \times k(q+1)}$ denote the augmented gallery dictionary, $\mathbf{V} \in \mathbb{R}^{d \times m}$ denote the variational dictionary, and $e$ is a noise term. This model searches for the sparsest representation of the probe sample in both $\mathbf{D^\prime}$ and $\mathbf{V}$ dictionaries. We first extend the original ESRC to the following robust formulation (Eq.~\ref{eq9}). 
\begin{equation}
\min_{\alpha,\beta} \bigg\|\mathbf{y}-\mathbf{[D^\prime, V]}
\begin{bmatrix}
\pmb{\alpha} \\ \pmb{\beta}
\end{bmatrix} \bigg\|_2^2  
+  \lambda  \big\| \pmb{\alpha} \big\|_1
+  \mu  \big\| \pmb{\beta}  \big\|_{\tau}, 
\label{eq9}
\end{equation}
where $\| \cdot \|_{\tau}$ corresponds with combination of Gaussian and Laplacian priors, defined as Eq. \ref{eq101}. This model assigns different regularization parameters to the $\pmb{\alpha}$ and $\pmb{\beta}$ coefficients to guaranty the robustness of the variational information from generic set~\cite{Li}.
\begin{equation}
\big\| {x} \big\|_{\tau} = \tau \big\| {x} \big\|_1 + (1 - {\tau}) \big\| {x} \big\|_2.
\label{eq101}
\end{equation}

The simultaneous sparsity constraint is then imposed to fully benefit from the variational information as well as synthetic still ROIs. 
Each generic set cluster found during the representative selection forms a block in the auxiliary dictionary, and exemplar of each cluster is considered as rendering parameter in face synthesizing for augmenting the gallery dictionary. The same sparsity pattern constraint in terms of the pose angle is imposed on the dictionaries which encourages similar pose angles to select the same set of atoms for representing each view. In this way, the coefficient vectors for the still ROIs in the augmented gallery dictionary are forced to share the same sparsity pattern with non-zero coefficients associated with the video ROI belonging to the corresponding block (cluster) of the same view in the auxiliary dictionary. This improves the discrimination power of the dictionaries accordingly. 
% The deep CNN features of both dictionaries are then extracted
The new sparse coefficients can be obtained by solving the following optimization problem: 
% \sum_{i=1}^n
\begin{equation}
\min_{{A},{B}} \bigg\|\mathbf{y} -\mathbf{[D^\prime, V]
\begin{bmatrix}
\mathbf{A} \\ \mathbf{B}
\end{bmatrix} }\bigg\|_F^2  
+  \lambda  \big\| \mathbf{A}  \big\|_1
+  \mu  \sum_{l=1}^{q}  \big\| \mathbf{B} [l] \big\|_{\tau},
\label{eq10}
\end{equation}
where $\| \cdot \|_F$ denotes the Frobenius norm, $\mathbf{A} = [\pmb{\alpha}_1,\pmb{\alpha}_2,\dotsc, \pmb{\alpha}_{k(q+1)}] $ and $\mathbf{B} = [\pmb{\beta}[1], \pmb{\beta}[2], \dotsc, \pmb{\beta}[q]]$ are coefficients matrix consists of $q$ blocks which $q$ is number of clusters/representatives.
% $ \| S \|_F^2 =  \sum_{k=1}^k \| S_k \|_2^2 $
\begin{equation}
\begin{bmatrix}
{\widehat{\mathbf{A}}} \\      
{\widehat{\mathbf{B}}}
\end{bmatrix} = {\arg\min} 
\bigg\| \mathbf{y}-\mathbf{[D^\prime, V]
\begin{bmatrix}
\mathbf{A} \\ \mathbf{B}
\end{bmatrix} 
}\bigg\|_F^2 +  \lambda  \big\| \mathbf{A}  \big\|_1
+  \mu  \sum_{l=1}^{q}  \big\| \mathbf{B} [l] \big\|_{\tau} ,
\label{eq11}
\end{equation}
subject to:
\begin{equation*}
\big\| 
\mathbf{A} , \mathbf{B}
\big\|_{2,1}  \leq \xi ,
\end{equation*}
where $\xi$ is the sparsity level and $\| \cdot \|_{2,1}$ is the mixed norm defined as the sum of $\ell_2-norm$ of all rows of matrix $\mathbf{A}$ and $\mathbf{B}$ and then applying $\ell_1-norm$ on the obtained vector.
Note that each view in formulation of Eq.~\ref{eq11} shares the same sparsity pattern at class-level, but not necessarily at atom-level in real world scenarios. This problem, called joint dynamic sparse representation, can be solved by applying $\ell_0-norm$ across the $\ell_2-norm$ of the dynamic active sets~\cite{Nasrabadi} as follows:
\begin{equation}
\begin{bmatrix}
\widehat{\mathbf{A}} \\      
\widehat{\mathbf{B}}
\end{bmatrix} = {\arg\min} 
\bigg\| \mathbf{y}-\mathbf{[D^\prime, V]
\begin{bmatrix}
\mathbf{A} \\ \mathbf{B}
\end{bmatrix} 
}\bigg\|_F^2 +  \lambda  \big\| \mathbf{A}  \big\|_1
+  \mu  \sum_{l=1}^{q}  \big\| \mathbf{B} [l] \big\|_{\tau},
\label{eq12}
\end{equation}
subject to:
\begin{equation*}
\big\| \mathbf{A} , \mathbf{B} \big\|_G  \leq \xi,
\end{equation*}
where $\big\| \cdot \big\|_G$ is defined as follows:
\begin{equation}
\big\| \mathbf{A} , \mathbf{B} \big\|_G = \bigg\| \begin{bmatrix}  \big\|\mathbf{A}_{g_{1}},\mathbf{B}_{g_{1}}\big\|_1, \big\|\mathbf{A}_{g{2}},\mathbf{B}_{g{2}} \big\|_2 , \dotsc \end{bmatrix} \bigg\|_0.
\end{equation}
where $x_{g_{i}}$ is a set coefficients associated with the $i^\textit{th}$ active set $g_i$
\begin{equation}
x_g = X(g_s(1),1), \dotsc, X(g_s(M),M) ]^T \in \mathbb{R}^{m}
\end{equation}
where $g_s$ for $s=1,2,\dotsc,k$ is dynamic active set refers to the indices of a set of coefficients belonging to the same class in the coefficient matrix. 
In order to solve this optimization problem, the classical alternating direction method of multipliers is considered \cite{tropp2006}.
The use of joint dynamic sparsity regularization term allows combining the cues from all the views during joint sparse representation. Moreover, it provides a better representation of the multiple view images, which represent different measurements of the same individual from different viewpoints.
Finally, the residuals for each class $k$ are calculated for the final
classification as follows:

\begin{equation}
%r_k(y)  =  \bigg\|\mathbf{\enskip y - [R^\prime_k, V_k]
r_k(y) =  \bigg\|\mathbf{y}-[\mathbf{D}^\prime, \mathbf{V}]
	\begin{bmatrix}
	\gamma_k (\widehat{\mathbf{A}}_k)\\ \widehat{\mathbf{B}}_k
	\end{bmatrix} 
	\bigg\|_F^2 ,
\label{eq8}
\end{equation}
where ${\gamma_{k}}$ is a vector whose nonzero entries are the entries in $\widehat{\mathbf{A}}_k$ that are associated with class $k$. Then the class with the minimum reconstruction error is regarded as the label for the probe subject $y$.
Algorithm \ref{alg1} summarizes the S+V model for still-to-video FR from a SSPP.
%\begin{equation}
%k^* = \underset{k}{\arg\min}  \| r_k(\mathbf{y}) \|_F^2 .
%\label{eq13}
%\end{equation}

%****************************************
\RestyleAlgo{boxruled}
\begin{algorithm}[ht]
	\caption{Synthetic Plus Variational Model.}
	\label{alg1}
	\SetAlgoLined
	\small
	\KwIn{Reference still ROIs $\mathbf{D} = \{ \mathbf{r}_1,\mathbf{r}_2, \dotsc ,\mathbf{r}_k \} \in \mathbb{R}^{d \times k}$, 
	Generic set $\mathbf{G} = \{ \mathbf{g}_1,\mathbf{g}_2, \dotsc ,\mathbf{g}_m \} \in \mathbb{R}^{d \times m}$, probe sample $\mathbf{y}$, and parameters $\lambda$, $\mu$, and $\xi$.} 
    	%\For {each $\mathbf{g}_{i}$ }{  
    Estimate pose angles of $\mathbf{G}$.\\
	Apply row sparsity clustering in the pose space of $\mathbf{G}$, and produce $q$ clusters (representative exemplars).\\
    Find center of each cluster as $q$ representative pose angles.\\  
    Construct the variation dictionary, $\mathbf{V} \in \mathbb{R}^{d \times m}$, with $q$ blocks.\\
	 \For {each $\mathbf{r}_{i}$ }{
    Generate $q$ synthetic images $\mathbf{S}_{i} \in \mathbb{R}^{d \times q}$ per each individual based on $q$ representative pose angle.\\
    Merge $\mathbf{S}_{i}$ with $\mathbf{r}_{i}$ to form $\mathbf{D}_{i}^\prime \in \mathbb{R}^{d \times (1+q)}$.\\
    %Extract deep features from $\mathbf{y}$, $\mathbf{D}_{i}^\prime$ and $\mathbf{V}_i$.\\
    }  
    %}
    Solve the sparse representation problem to estimate coefficient matrix, $\mathbf{A}$ and $\mathbf{B}$, for $y$ by Eq.~\ref{eq12}.\\
	Compute the residual, $r_k(y)$ by Eq.~\ref{eq8}.\\                
	\KwOut{$label(y) = \underset{k}{\arg\min} (  r_k (y) ) $.}
\end{algorithm}

%********************************************************************
\section{Still-to-Video Face Recognition with the S+V Model} 
\label{FR}
\begin{figure}[ht]
\centering 
\includegraphics[width=.75\textwidth]{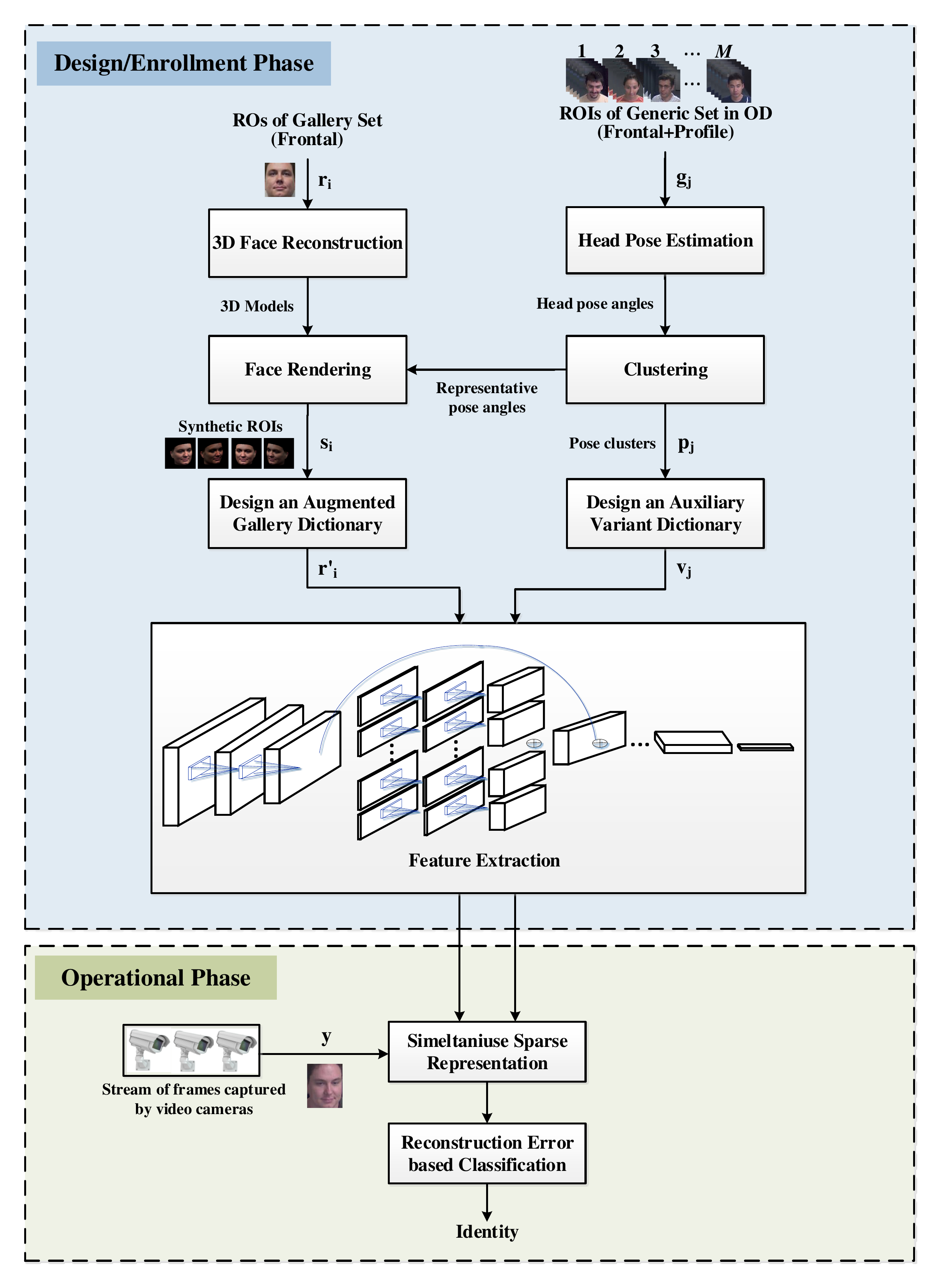}
\caption{\small Block diagram of the proposed still-to-video FR system with the S+V modeling.}
\label{fig4}
\end{figure}

%********************************************************************
In this section, a particular implementation is considered (see Fig.~\ref{fig4}) to assess the impact of using the S+V model for still-to-video FR. 
The augmented and auxiliary dictionaries are constructed by employing the representative synthetic ROIs and generic variations, respectively, and classification is performed by SRC while the generic set in the auxiliary dictionary is forced to combine with approximately the same facial viewpoint in the augmented gallery dictionary. The main steps of the proposed domain-invariant FR with the S+V model are summarized as follows.

\begin{itemize}
\item {\textbf{Step 1.} Select Representatives:}
The generic set $\mathbf{G}_i \in R^{{d}\times{m}}$ in the target domain is clustered based on their pose angles based on row sparsity.
%-----------
\item {\textbf{Step 2.} Design an Augmented Gallery Dictionary:}
The $q$ synthetic ROIs $\mathbf{S}_i \in R^{{d}\times{q}}$ are generated for each $\mathbf r_i$ of the reference gallery set in the source domain to form an augmented gallery dictionary $\mathbf{D}_i^\prime \in \mathbb{R}^{d \times k(q+1)}$, where $q$ is the number of clusters/representatives. 
%-----------
\item {\textbf{Step 3.} Form an Auxiliary Dictionary:}
The variations of the natural albedo of the generic set $\mathbf{G}_i \in R^{{d}\times{m}}$ in the target domain are extracted by subtracting the natural image from other images of the same class to form a generic auxiliary dictionary $\mathbf{V}_i \in R^{{d}\times{m}}$ with block structure.
%-----------
\item {\textbf{Step 4.} Extract Features:}
The deep CNN features of $\mathbf{D}_i^\prime \in \mathbb{R}^{d \times k(q+1)} $ and $\mathbf{V}_i \in R^{{d}\times{m}}$ are extracted.
%-----------
\item {\textbf{Step 5.} Apply Simultaneous Sparsity:}
The augmented gallery dictionary is encouraged to pair the sparsity pattern with the auxiliary dictionary for the same pose angles by applying the simultaneous sparsity.
%-----------
\item {\textbf{Step 6.} Validation:} 
The proposed system assess if given probe ROIs belong to one of the enrolled persons and rejects invalid probe ROIs using \textit{sparsity concentration index (SCI)} criteria defined in~\cite{Wright1}:

\begin{equation}
\mbox{SCI}({\hat{\alpha}}) \doteq \frac{k.{\underset{i}{\max}}  \parallel {\delta_i} ({\hat{\alpha}})\parallel_1/\parallel {\hat{\alpha}} \parallel_1 - 1 }{k-1} \quad \in [0,1]
\label{eq20}~.
\end{equation}
%
% where $k$ is the number of classes. 
A probe ROI is accepted as valid if SCI$({\hat{\alpha}})\geq\tau$ and otherwise rejected as invalid, where $\tau \in (0,1)$ is an outlier rejection threshold. 
\end{itemize}

%**************************************************
\section{Experimental Methodology} 
\label{experiment}
%**************************************************

%************************************
\subsection{Datasets:}
In order to evaluate the performance of the proposed S+V model for still-to-video FR, an extensive series of experiments are conducted on Chokepoint\footnote{{\href{http://arma.sourceforge.net/chokepoint/}{http://arma.sourceforge.net/chokepoint}.}}~\cite{Wong} and COX-S2V\footnote{{\href{http://vipl.ict.ac.cn/view_database.php?id=3/}{http://vipl.ict.ac.cn}.}}~\cite{Huang3} datasets. Chokepoint~\cite{Wong} and COX-S2V~\cite{Huang3} datasets are suitable for experiments in still-to-video FR in video surveillance because they are composed of a high-quality still image and lower-resolution video sequences, with variations of illumination conditions, pose, expression, blur and scale. 

\begin{figure}[ht]
        \centering
        \subfigure[Still Reference ROIs]{\label{fig:a}  
        \includegraphics[width=.5\textwidth]{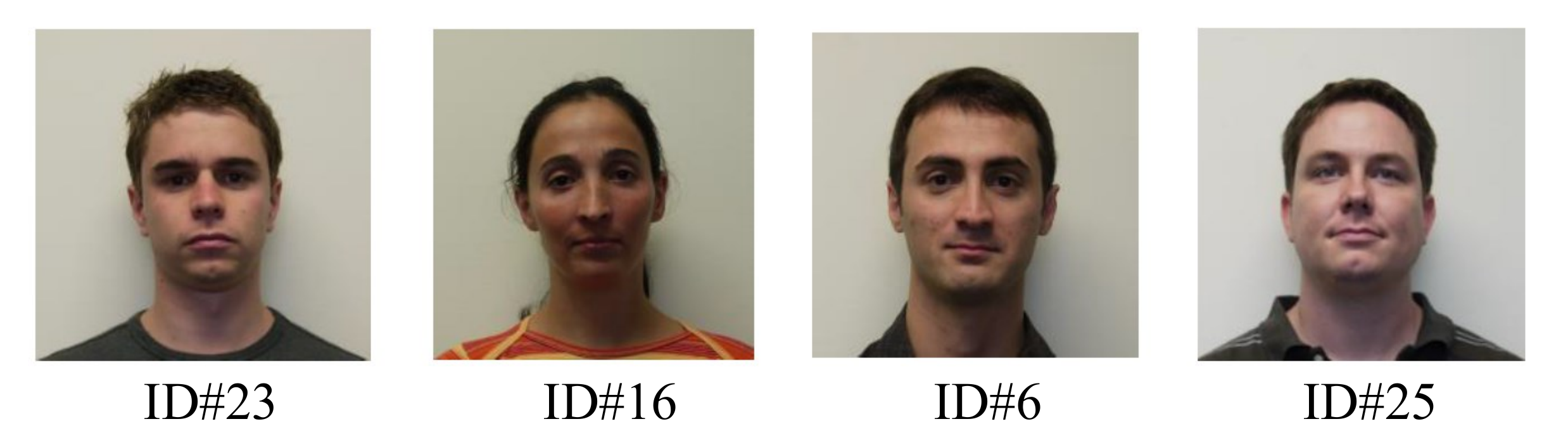}}
    ~ 
        \centering
        \subfigure[Examples of Video ROIs]{\label{fig:a}
        \includegraphics[width=.58\textwidth]{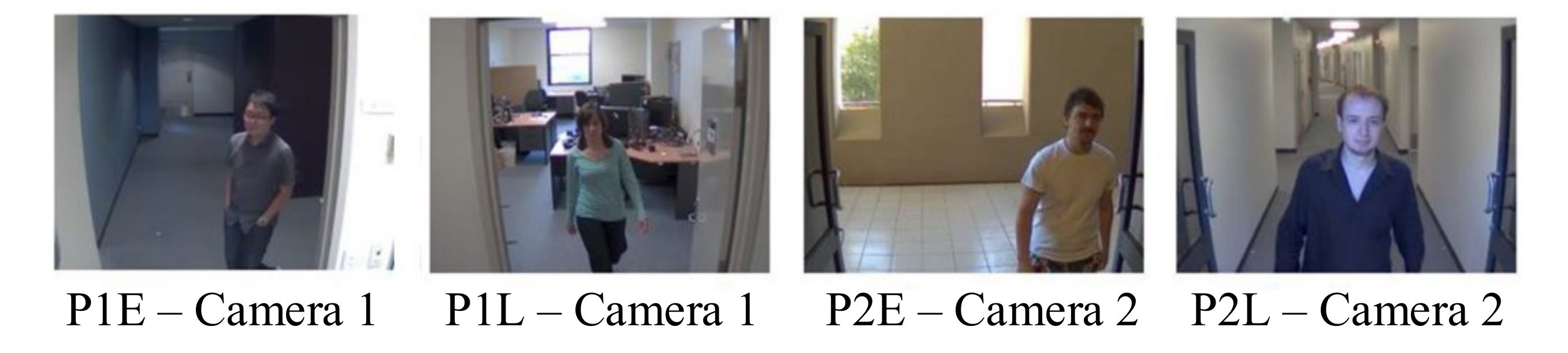}}
        \vspace{-.1cm}
        \caption{\small Examples of still images and video frames from portals and cameras of Chokepoint dataset.}
         \vspace{-.1cm}
        \label{fig5} 
\end{figure}

Chokepoint~\cite{Wong} (see Fig.~\ref{fig5}) consists of $25$ subjects walking through portal $1$ (P1) and $29$ subjects in portal $2$ (P2). Videos are recorded over $4$ sessions (S1,S2,S3,S4) one month apart. An array of $3$ cameras (Cam1,Cam2,Cam3) are mounted above P1 and P2 that capture the subjects during $4$ sessions while they are either entering (E) or leaving (L) the portals in a natural manner. In total, $4$ data subsets are available (P1E, P1L, P2E, and P2L), and the dataset consists of $54$ video sequences. 

\begin{figure}[ht]
        \centering
        \subfigure[Still Reference ROIs]{\label{fig:a}  
        \includegraphics[width=.51\textwidth]{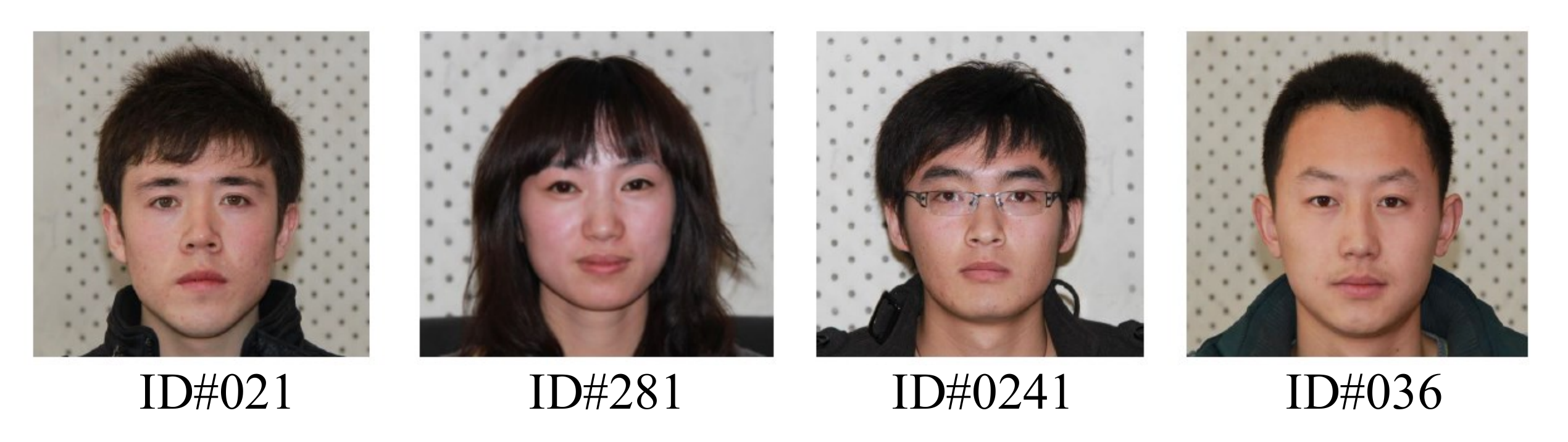}}
    ~ 
        \centering
        \subfigure[Examples of Video ROIs]{\label{fig:a}
        \includegraphics[width=.62\textwidth]{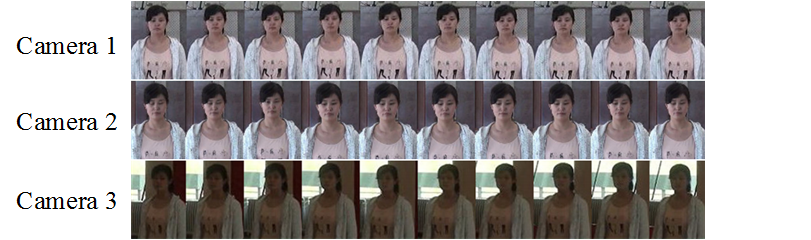}}
        \vspace{-.1cm}
        \caption{\small Examples of still images and video frames from  3 cameras of COX-S2V dataset.}
        \label{fig6} 
\end{figure}

COX-S2V dataset~\cite{Huang3} (see Fig.~\ref{fig6}) contains $1,000$ individuals, with $1$ high-quality still image and $3,000$ low-resolution video sequences per each individual simulating video surveillance scenario. The video frames are captured by $4$ cameras (Cam1, Cam2, Cam3, Cam4) mounted at fixed locations of about $2$ meters high. In each video, an individual walk through an S-shape route with changes in pose, illumination, and scale. 

%**************************

%************************************
\subsection{Protocol and Performance Measures:} 

A particular implementation of the S+V model for still-to-video FR has been considered to validate the proposed approach. We  hypothesize that accuracy can be improved by adding synthetic reference faces to the gallery dictionary and encouraging the dictionaries to share the same sparsity pattern for the same pose angles can address non-linear pose  variations. 

First, it is assumed that during the calibration process, $q$ representative pose angles are selected based on the $q$ pose clusters obtained from facial ROI trajectories of unknown persons captured in the target domain using the row sparsity clustering. 
During the enrollment of an individual to the system, $q$ synthetic ROIs for each reference still ROI are generated under typical pose variations from different camera viewpoints. For face synthesis, we employ the conventional 3D Morphable Model (3DMM)~\cite{blanz1} and the CNN-regressed 3DMM~\cite{Tran1}, that relies on a CNN for regressing 3DMM parameters. The gallery dictionary is constructed using the reference still ROIs of the individuals along with their synthetic ROIs. Next, the auxiliary variational dictionary is designed using the intra-class variations of the generic set with block structure ($q$ blocks). 
Additionally, we consider extracting deep features using CNN models to further improve the FR recognition rate. The networks are pre-trained using the VGGFace2 dataset with AlexNet~\cite{krizhevsky}, ResNet~\cite{He} and VGGNet~\cite{simonyan} architectures using Triplet Loss \cite{facenet}.
The extracted features are concatenated as a row feature vector of this dictionary. The sparse model is fed with the extracted features. In all experiments with Chokepoint dataset, $5$ target individuals are selected randomly to design a watch-list that includes a high-quality frontal captured images, and for the experiment with COX-S2V, $20$ individuals are randomly selected to build a watch-list from high-quality faces. Videos of $10$ individuals that are assumed to come from non-target persons are used as generic set. The rest of the videos including $10$ other non-target individuals and $5$ videos of individuals who are already enrolled in the watch-list are used for testing. In order to obtain representative results, this process is repeated $5$ times with a different random selection of watch-lists and the average performance is reported with standard deviation over all the runs.

During the operational phase, FR is performed by sparse coding the features of probe ROI over the features of augmented and auxiliary (variational) dictionaries ROIs.  The sparsity parameter $\Lambda$ is fixed to $0.005$ during the experiments. We also compared the S+V method to several baseline state-of-the-art methods: {ESRC}~\cite{Deng}, {SVDL}~\cite{Yang}, RADL~\cite{Wei}, {LGR}~\cite{Zhu}, {CSR}~\cite{Li}, face frontalization~\cite{Hassner}, and recognition via generation~\cite{masi2}.

The average performance of the proposed and baseline FR systems is measured in terms of accuracy and complexity. For accuracy, we measure the partial area under ROC curve pAUC(20\%) (using the AUC at $0 < FPR \leq 20$\%) and area under precision-recall space (AUPR).  An estimation of time complexity is provided analytically based on the worst-case number of operations performed per iteration. Then, the average running time of our algorithm is measured with a randomly selected probe ROIs using a PC workstation with an Intel Core i7 CPU (3.41GHz) processor and 16GB RAM.

%********************************************
%********************************************
\section{Results and Discussion} 
\label{results}
%*******************************************

This section first shows some examples of synthetic face images produced under representative pose variations, and then presents still-to-video FR performance achieved when augmenting SRC dictionaries with such images to address non-linear variations caused by pose changes. In order to investigate the impact of the proposed S+V model on performance, we considered the still-to-video FR system described in Section~\ref{FR} with a growing number of synthetic faces, along with a generic training set. Finally, this section presents an ablation study (showing the effect of each module on the performance) and a complexity analysis for our proposed approach.

%**********************************
\subsection{Synthetic Face Generation:}
\begin{figure}[!t]
        \centering
        \subfigure[Chokepoint]{\label{fig:a} 
        \includegraphics[width=58mm]{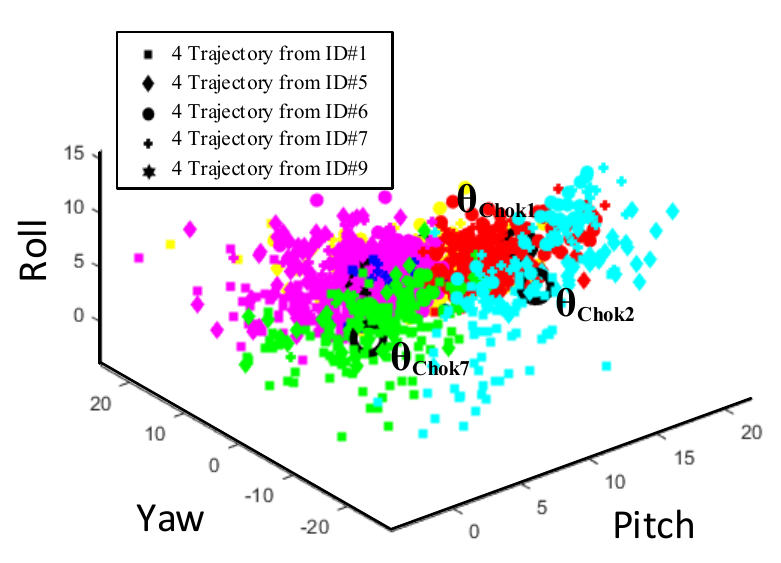}}
    ~ 
        \centering
        \subfigure[COX-S2V]{\label{fig:b}
        \includegraphics[width=58mm]{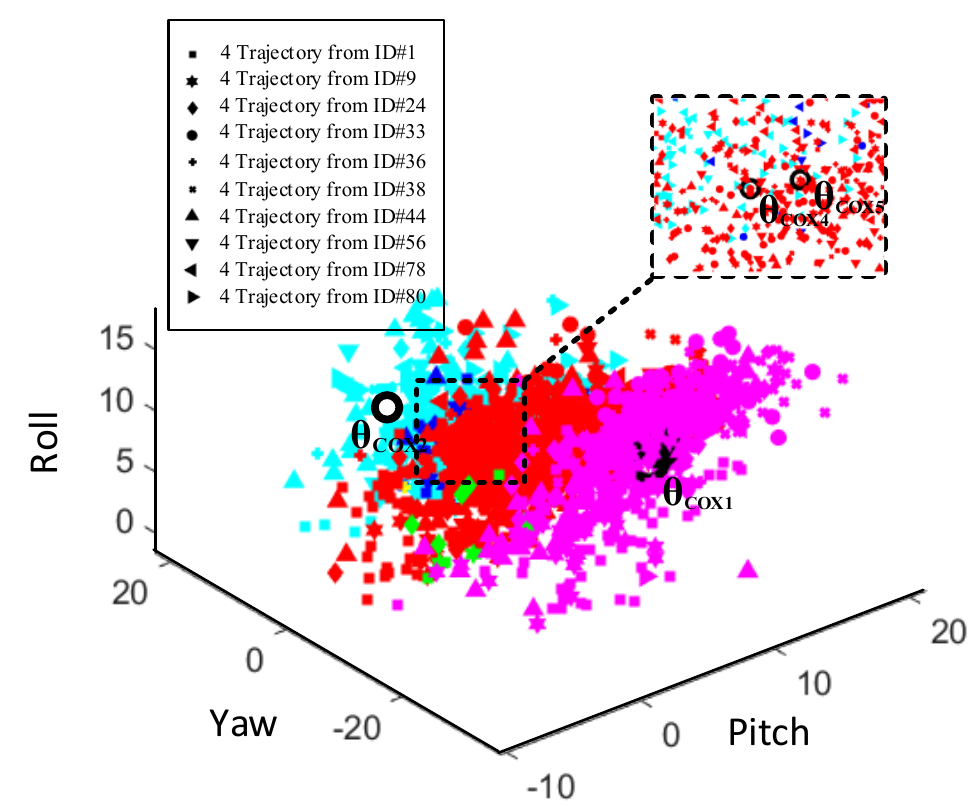}}
        \vspace{.2cm}
        \caption{\small Example of clusters obtained with $20$ and $40$ facial trajectories represented in the pose space with Chokepoint (ID\#1, \#5, \#6, \#7, \#9) and COX-S2V (ID\#1, \#9, \#24, \#33, \#36, \#38, \#44, \#56, \#78, \#80) datasets, respectively. Clusters are shown with different colors, and their representative pose exemplars are indicated with a black circle.}
        \label{fig7} 
\end{figure}

Fig.~\ref{fig7} shows an example of the clustering (based on row sparsity) obtained with facial ROIs of $20$ trajectories extracted from Chokepoint videos of $5$ individuals and $40$ trajectories extracted from COX-S2V videos of $10$ individuals in the 3-dimensional pose (roll-pitch-yaw) space. In this experiment, $q_{Chok}=7$ and $q_{COX}=6$ representative pose condition clusters are typically determined using row sparsity with Chokepoint and COX-S2V data, respectively. The exemplars selected from these clusters (black circles) are used to define representative pose angles for synthetic face generation with 3DMM and 3DMM-CNN techniques. For instance, the representative pose angles with the Chokepoint database, are listed as follows: ${\theta_{Chok}}_1 =$ (pitch, yaw, roll)= (15.65, 14.77, -0.62),  ${\theta_{Chok}}_2 =$ (12.44, 2.76, 3.64), ${\theta_{Chok}}_3=$ (9.06, -5.46, 4.73), ${\theta_{Chok}}_4 =$ (1.98,  6.09, 2.79), ${\theta_{Chok}}_5 =$ (13.21, 15.32, 6.14), ${\theta_{Chok}}_6=$ (0.64, -18.93, 0.86), ${\theta_{Chok}}_7=$ (5.23, 2.92, 2.03) degrees.

\begin{figure}[!t]
%\vspace{-1cm}
   \centering
    \includegraphics[width=.79\textwidth]{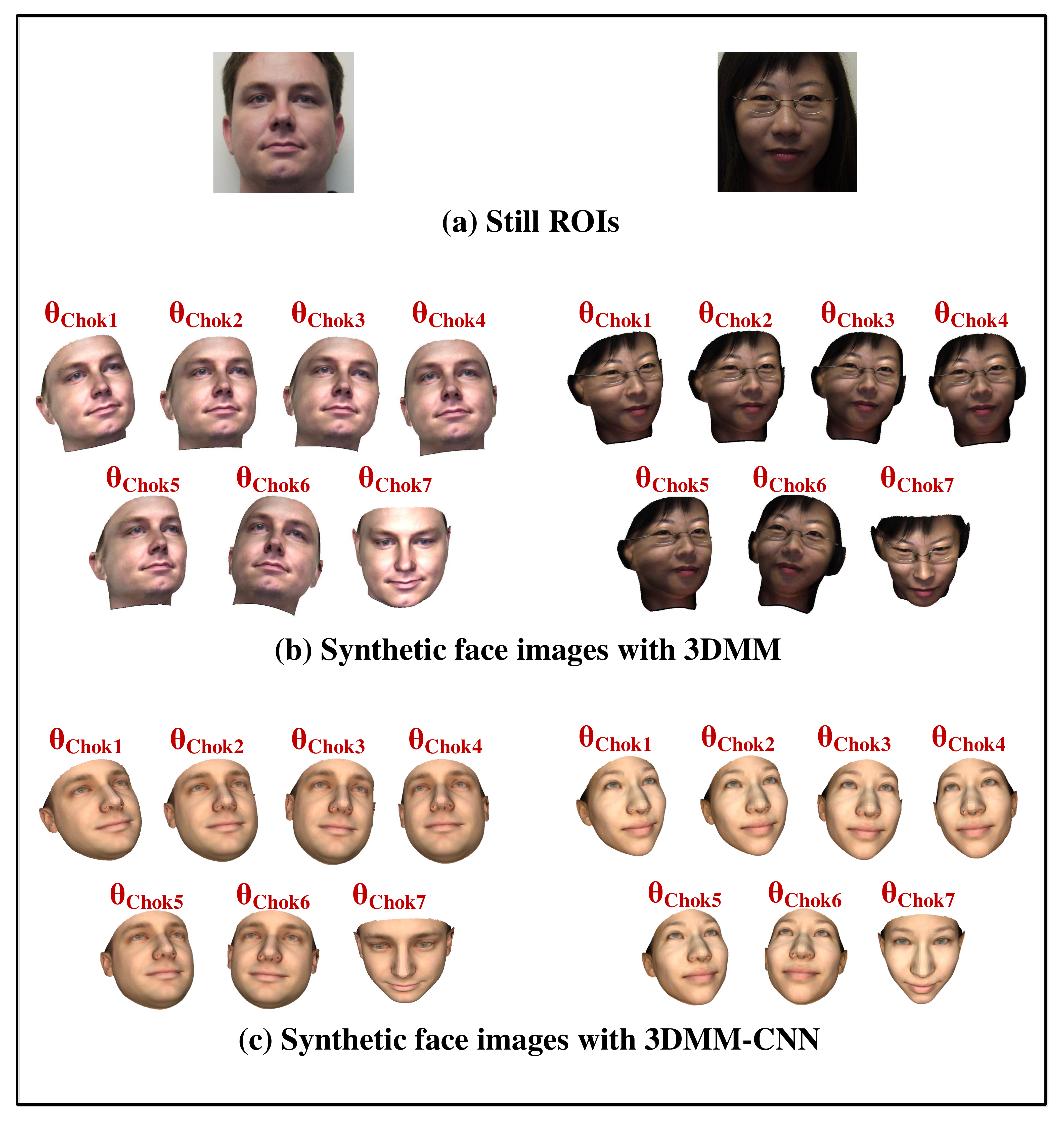}   
    \vspace{0.3cm}
	\caption{\small Examples of synthetic face images generated from the reference still ROI of individuals ID\#25 and ID\#26 (a) of Chokepoint dataset. They are produced based on representative exemplars (poses) and using 3DMM (b) and 3DMM-CNN (c).} 
	\label{fig8}
\end{figure}

\begin{figure}[!t]
   \centering
    \includegraphics[width=.70\textwidth]{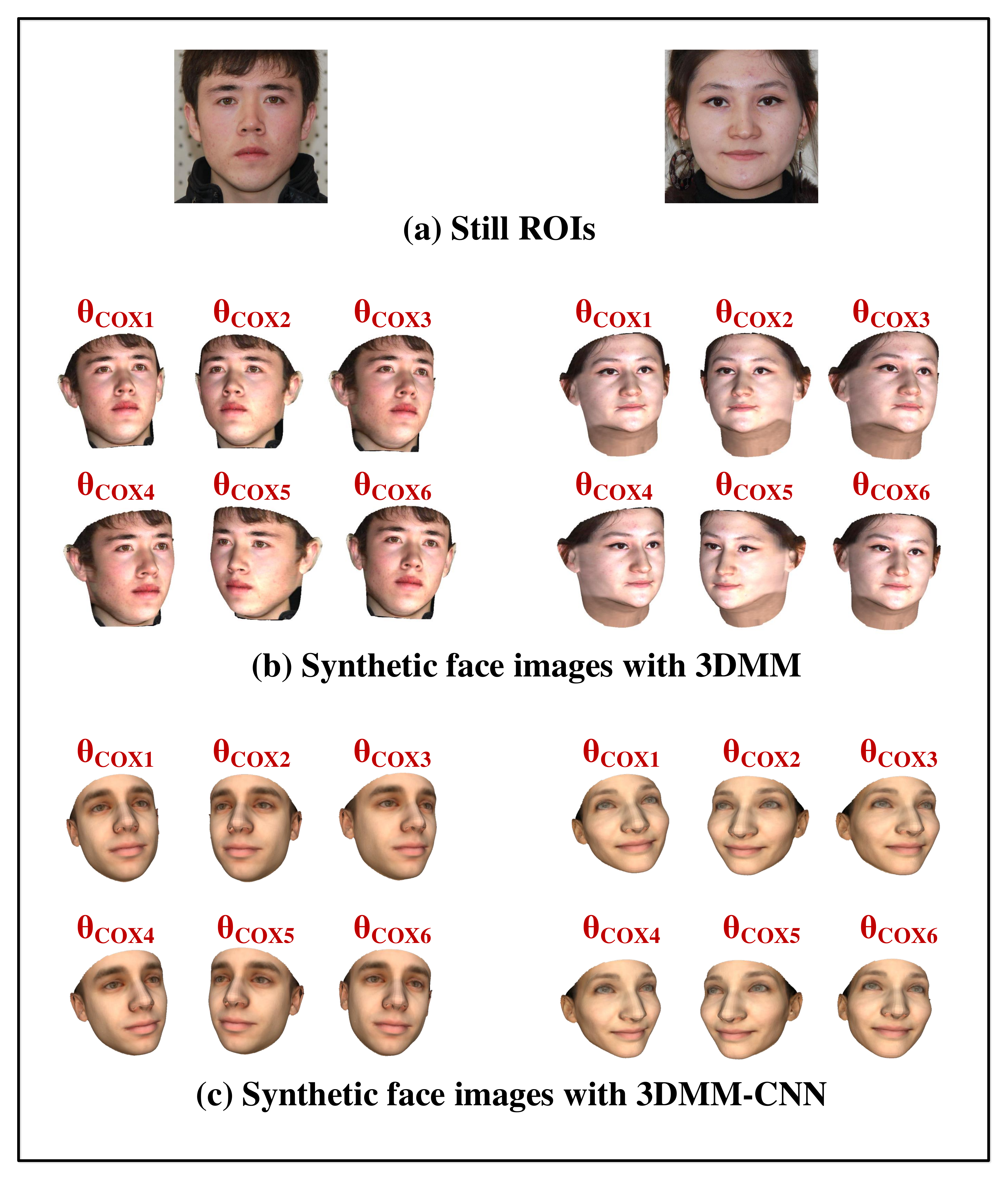}   
    \vspace{0.3cm}
	\caption{\small Examples of synthetic face images generated from the reference still ROI of individuals ID\#21 and ID\#151 (a) of COX-S2V dataset. They are produced based on representative exemplars (poses) and using 3DMM (b) and 3DMM-CNN (c).} 
	\label{fig9}
\end{figure}

Figs.~\ref{fig8} and~\ref{fig9} show the synthetic face images generated based on 3DMM and 3DMM-CNN under representative exemplars using reference still ROIs of the Chokepoint and COX-S2V datasets, respectively.   

%\begin{figure}[ht]
%    \centering
%    \subfigure[Chokepoint]{\label{fig:a}\includegraphics[width=60mm]{fig/IM2}}
%    ~ 
%    \centering
%    \subfigure[Chokepoint]{\label{fig:a}\includegraphics[width=55mm]{fig/IM3}}
%    \vspace{-.1cm}
%    \caption{\small Average pAUC and AUPR accuracy versus the size of the synthetic set on Chokepoint (a,b) and COX-S2V (c,d) databases. } \label{fig8} 
%\end{figure}

%***********************************************
\subsection{Impact of Number of Synthetic Images:}

\begin{figure}[ht]
    \centering
    \subfigure[Chokepoint]{\label{fig:a}\includegraphics[width=58mm]{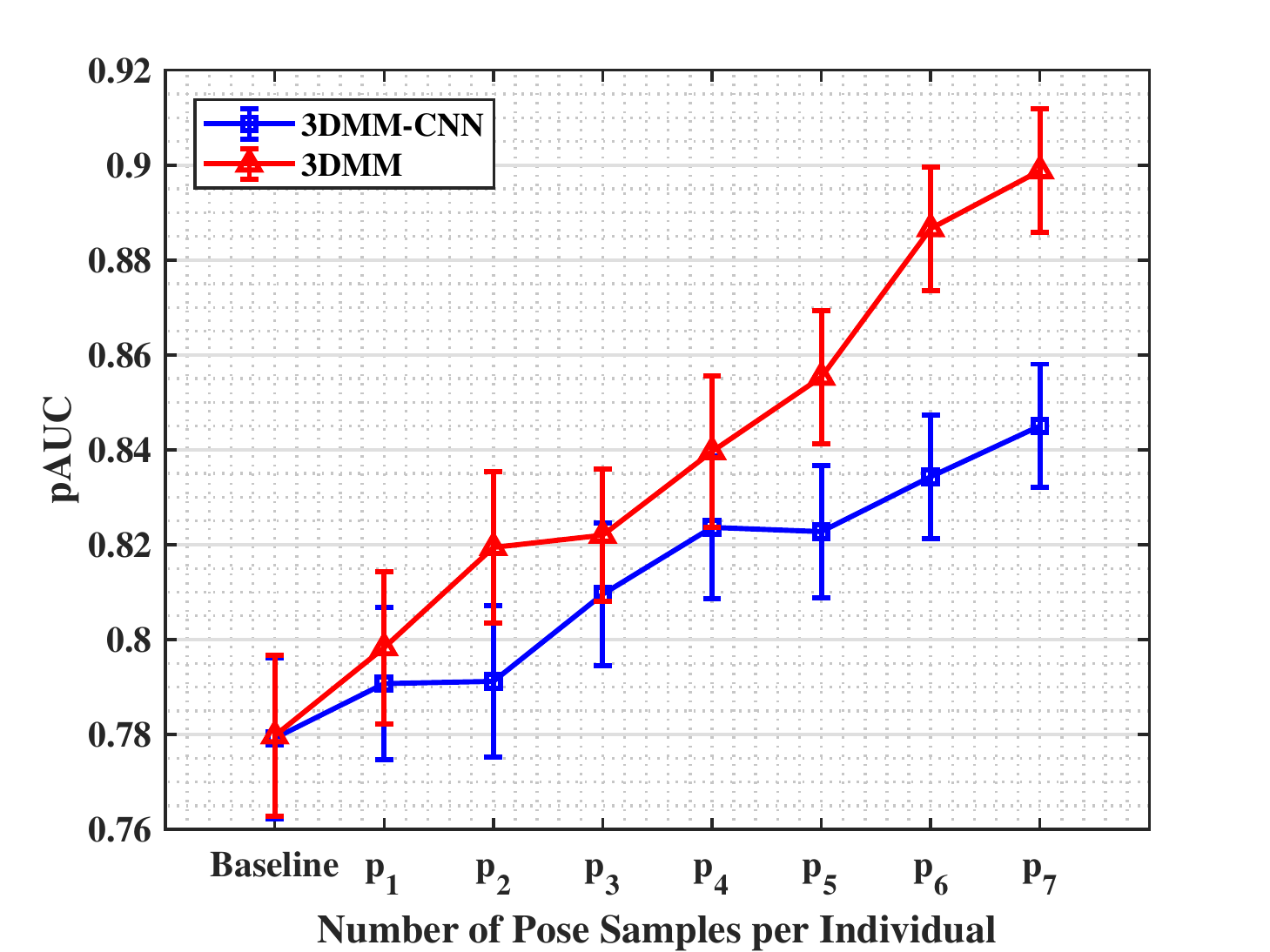}}
    ~ 
    \centering
    \subfigure[Chokepoint]{\label{fig:a}\includegraphics[width=58mm]{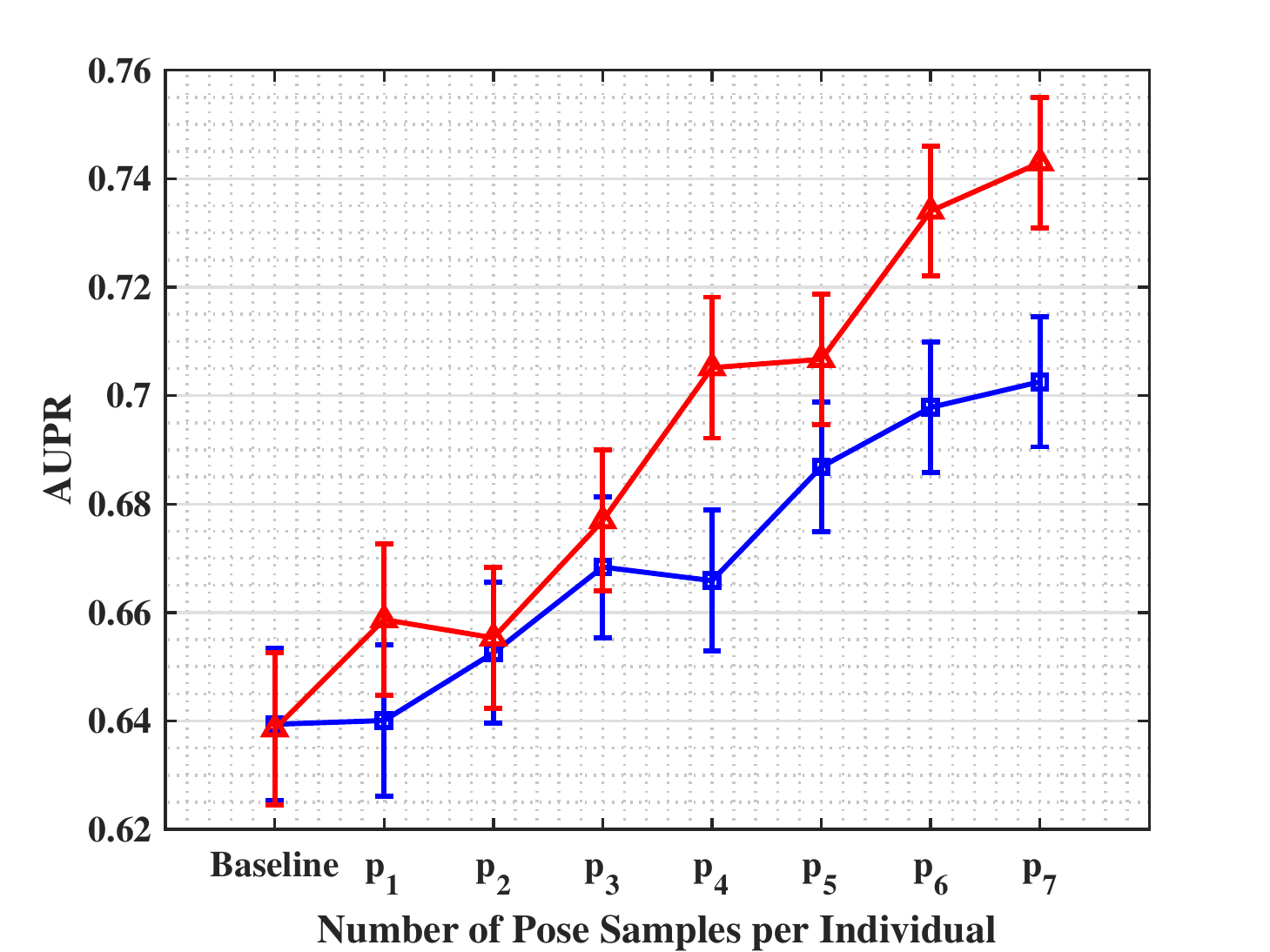}}
    ~
    \centering
    \subfigure[COX-S2V]{\label{fig:a}\includegraphics[width=58mm]{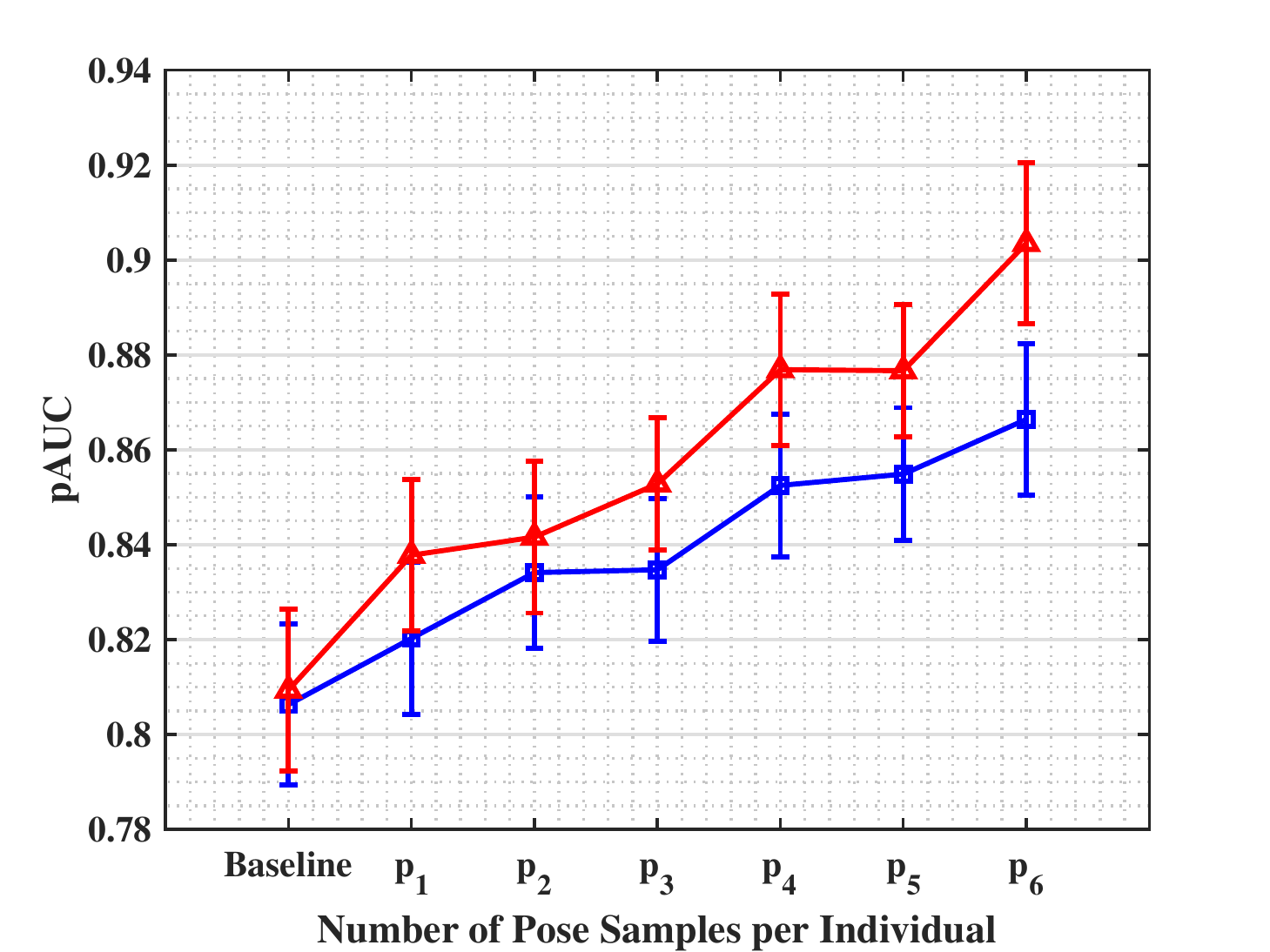}}
    ~ 
    \centering
    \subfigure[COX-S2V]{\label{fig:a}\includegraphics[width=58mm]{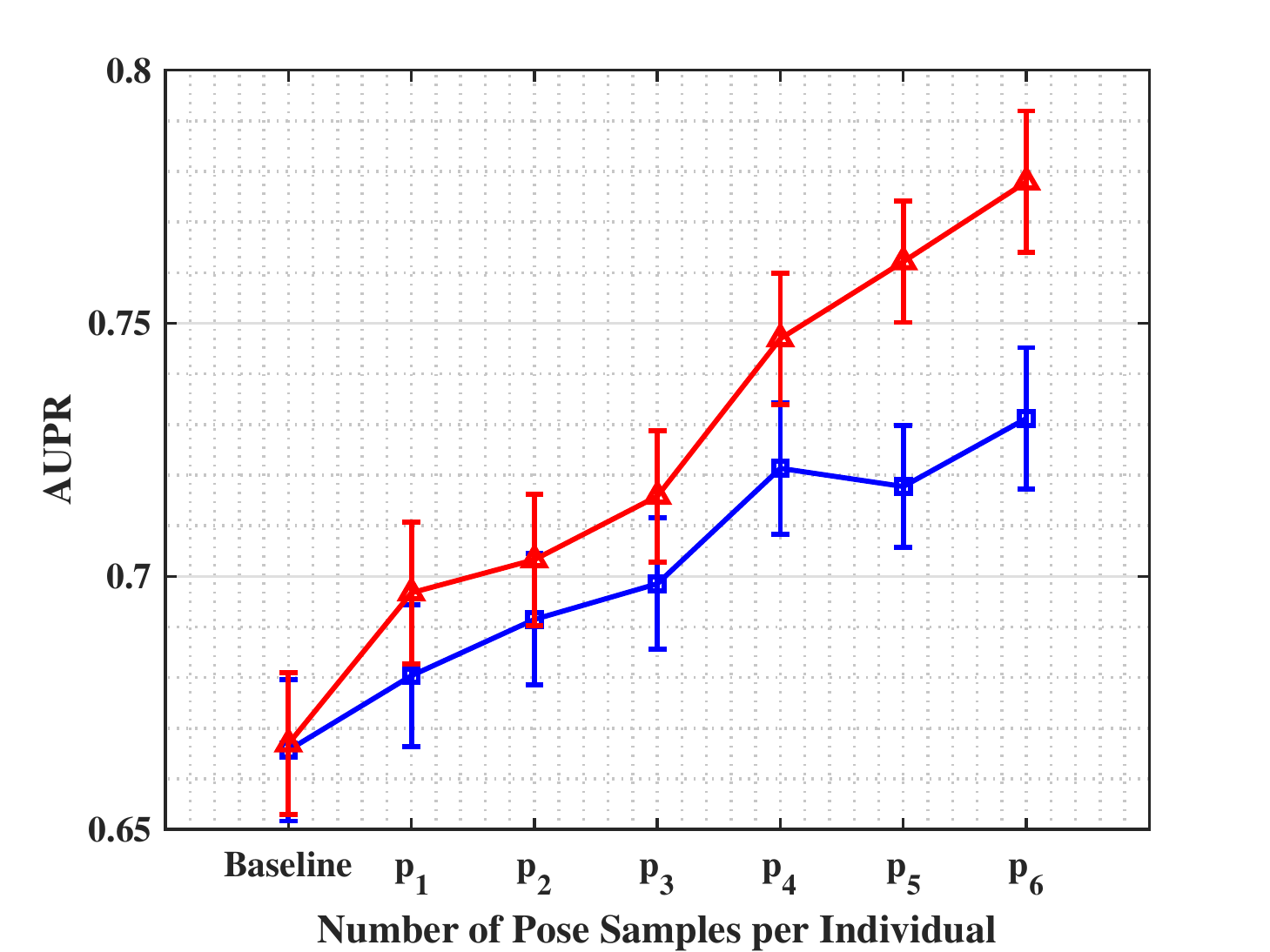}}
    \caption{\small Average pAUC(20\%) and AUPR accuracy of S+V model versus the size of the synthetic set generated using 3DMM and 3DMM+CNN on Chokepoint (a,b) and COX-S2V (c,d) databases. Error bars are standard deviation.} 
    \label{fig10} 
\end{figure}

In this subsection, the proposed S+V model is evaluated for a growing set of synthetic facial images in the augmented gallery dictionary. Fig.~\ref{fig10} shows the average pAUC(20\%) and AUPR accuracy obtained for the implementation in Section~\ref{FR} when increasing the number of synthetic ROIs per each individual. These ROIs were sampled from the $q$ representative pose exemplars from the Chokepoint and COX-S2V datasets. Results indicate that adding representative synthetic ROIs to the gallery dictionary allows to outperform the baseline system designed with an original reference still ROI alone. AUC and AUPR accuracy increase considerably by about $20 - 30\%$ with only $q_{Chok}=7$ and $q_{COX}=6$ synthetic pose ROIs (1 sample per pose cluster) for Chokepoint and COX-S2V datasets, respectively.

\begin{figure}[ht]
    \centering
    \subfigure[Chokepoint]{\label{fig:a}\includegraphics[width=58mm]{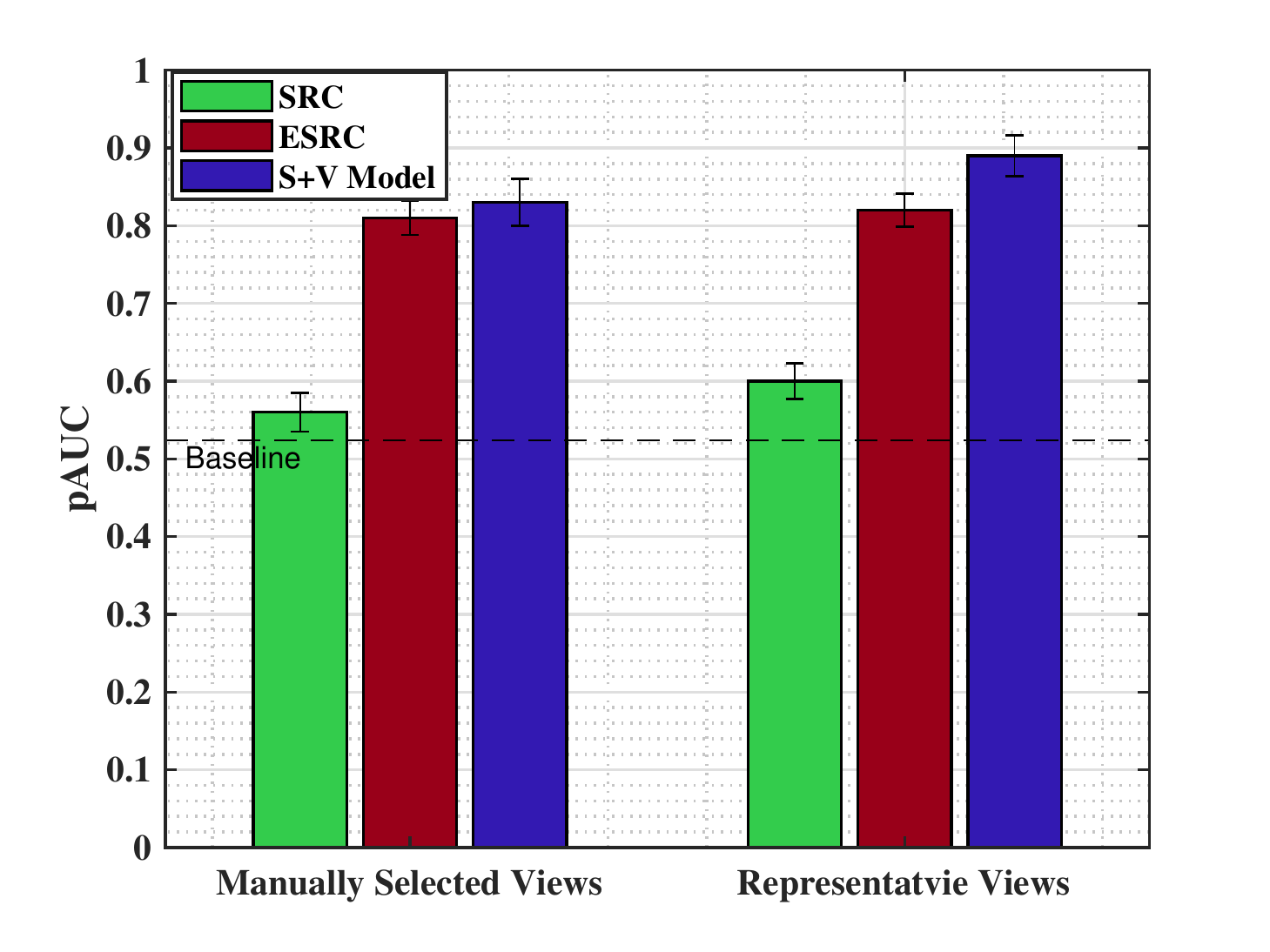}}
    ~ 
    \centering
    \subfigure[Chokepoint]{\label{fig:a}\includegraphics[width=58mm]{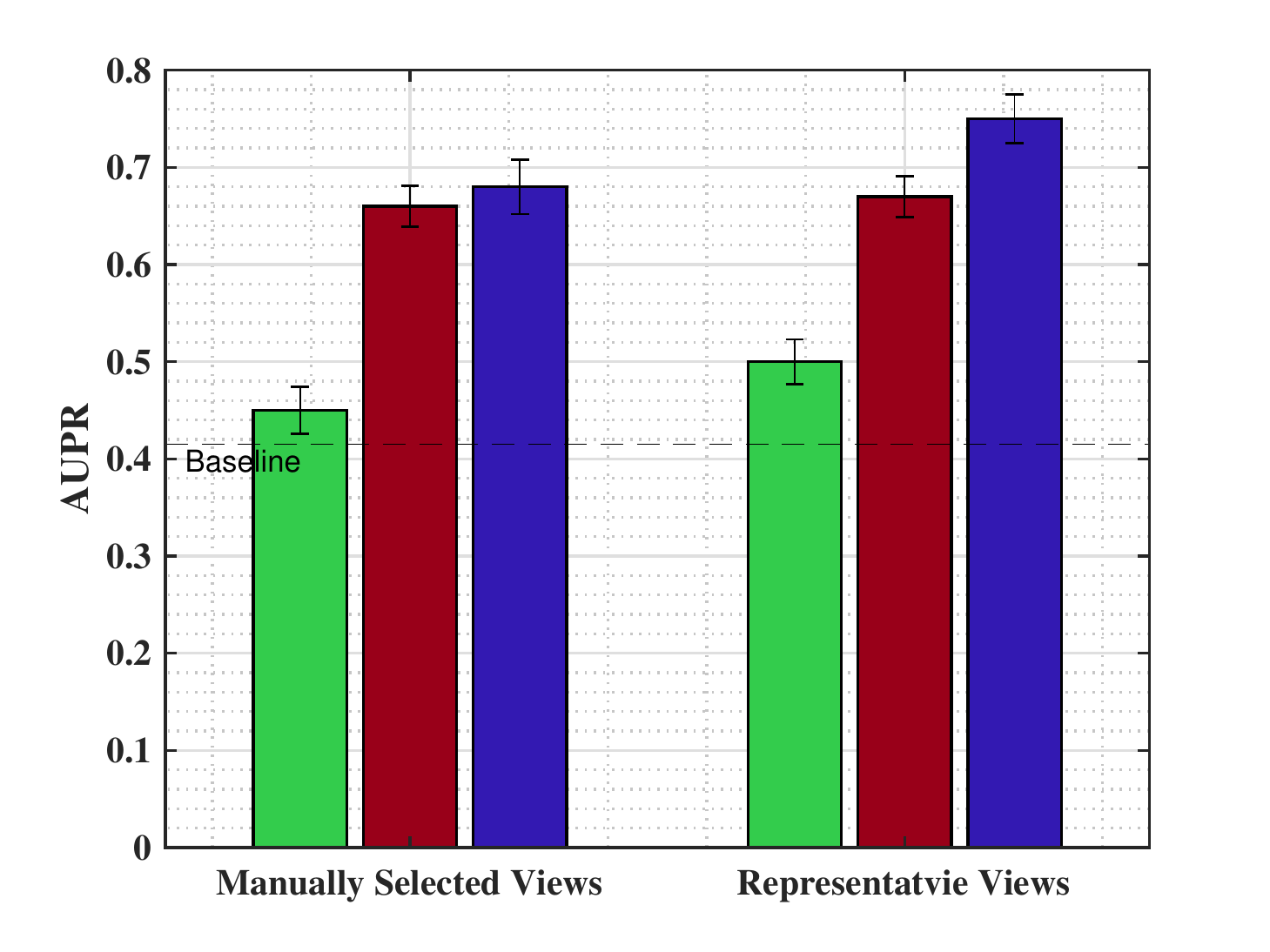}}
    ~
    \centering
    \subfigure[COX-S2V]{\label{fig:a}\includegraphics[width=58mm]{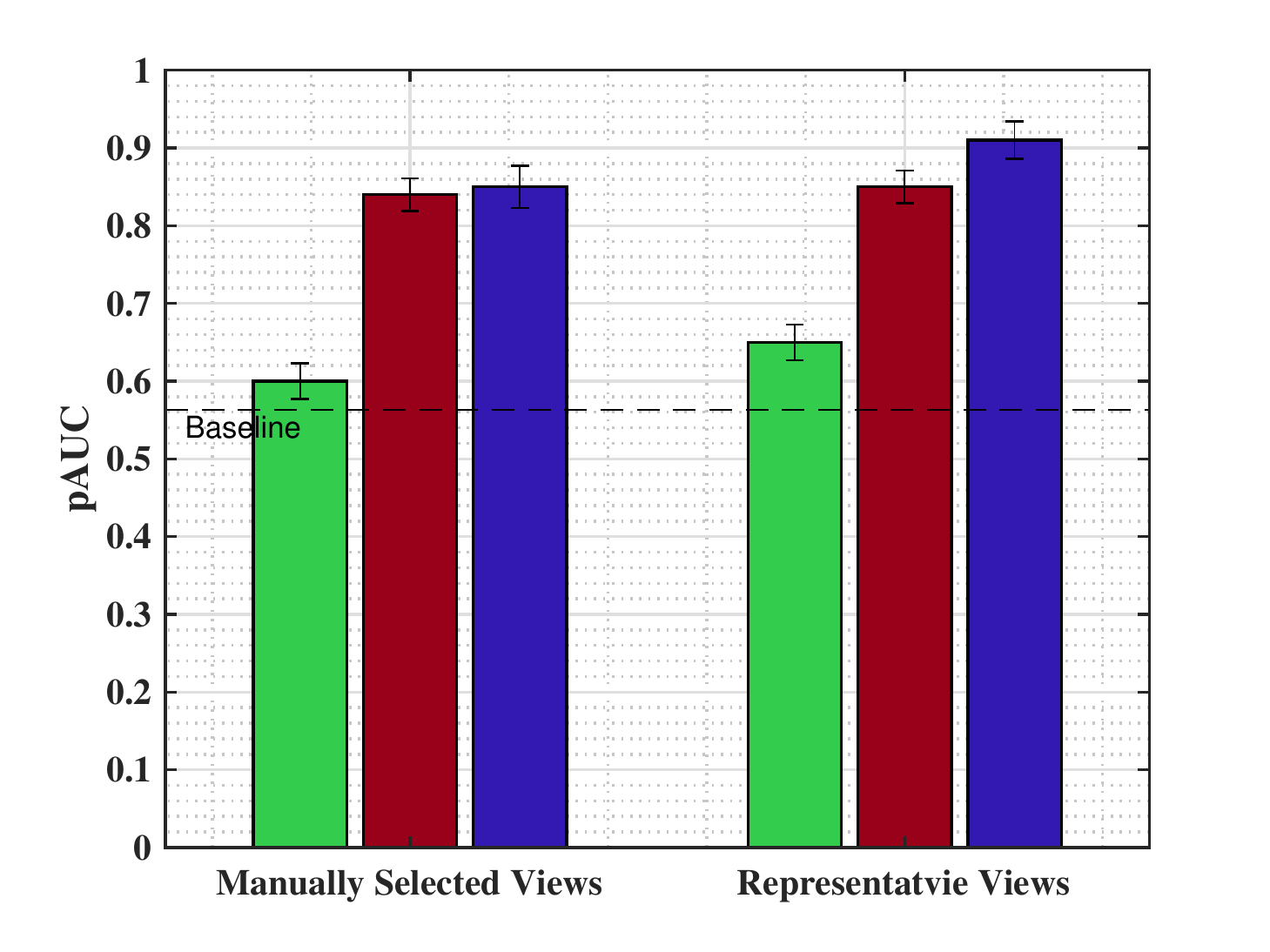}}
    ~ 
    \centering
    \subfigure[COX-S2V]{\label{fig:a}\includegraphics[width=58mm]{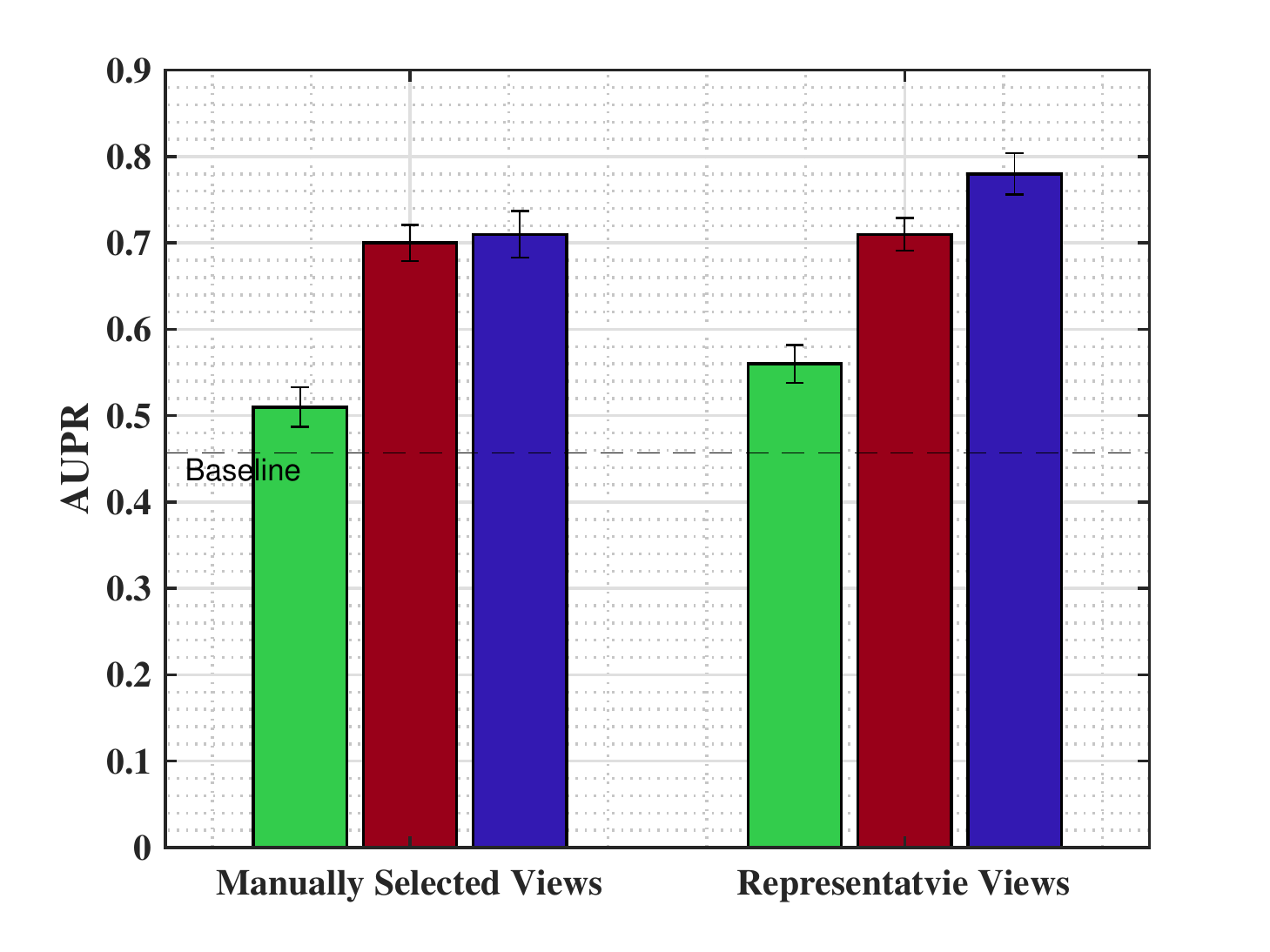}}
    \caption{\small Average pAUC(20\%) and AUPR accuracy for SRC, ESRC and S+V model on Chokepoint (a,b) and COX-S2V (c,d) databases. Error bars are standard deviation. } \label{fig11} 
\end{figure}

To further assess the benefits, Fig.~\ref{fig11} compares the performance of the proposed S+V method (adds $q$ synthetic samples) with the original SRC (without an auxiliary dictionary), and to ESRC (with manually designed auxiliary dictionary). Results in this figure show that the proposed method outperforms the others, and that FR performance is higher when the dictionary is designed using the representative views than based on the manually designed dictionary. The proposed method can therefore adequately generate representative facial ROIs for the gallery, and then match it with the corresponding variations in the auxiliary dictionary. Encouraging pair-wise relationships between the variational and augmented gallery dictionaries has a positive impact on the performance of still-to-video FR system based on SRC.

%*******************************************************
\subsection{Impact of Camera Viewpoint:}

To evaluate the robustness of the proposed S+V model to pose variations, accuracy is measured for different portals and video cameras, as well as for a fusion of cameras. Tables~\ref{table1} and \ref{table2} summarize the average accuracy on Chokepoint and COX-S2V datasets, respectively. For the Chokepoint dataset, videos are captured over 4 sessions for $3$ cameras (Camera1, Camera2, Camera3) over portals 1 (P1E, P1L) and portal 2 (P2E, P2L), while for the COX-S2V dataset, videos are captured over $3$ cameras (Camera1, Camera2 and Camera3). The performance of the S+V model is  compared with that of SRC and ESRC using the same configurations. Results show that the S+V model outperforms other techniques across different pose variations. Using synthetic profile views can improve the robustness of FR systems to pose variations. As expected, designing a system that combines faces from all the cameras (and portals) always provides a higher level of accuracy. 

\begin{table}[h!]
\vspace{.6cm}
\caption{\small Average accuracy of FR systems based on the proposed S+V model, SRC, and ESRC over different sessions, portals and cameras of the Chokepoint dataset. Feature representations are raw pixels, the 3DMM method is used for face synthesis.}
\label{table1}
\centering
\begin{adjustbox}{width=1\textwidth}
\begin{tabular}{clc|cccccc}
\hline
\multicolumn{2}{c}{\multirow{3}{*}{\textbf{Portal}}}    & \multirow{3}{*}{\textbf{Viewpoint}} & \multicolumn{5}{c}{\textbf{Accuracy}}     \\ 
               & & &  \multicolumn{2}{c}{\textbf{SRC}}  & \multicolumn{2}{c}{\textbf{ESRC}} &  \multicolumn{2}{c}{\textbf{S+V Model}}  \\ \cline{4-9} 
\multicolumn{2}{c}{}                          &         &    pAUC(20\%)   &       AUPR           &    pAUC(20\%)     &     AUPR        &  pAUC(20\%)   &     AUPR    \\ \hline \hline
\multicolumn{2}{c}{\multirow{4}{*}{P1}}& Camera1 & 0.482$\pm$0.023 & 0.361$\pm$0.021 &    0.691$\pm$0.020 & 0.534$\pm$0.023 &   0.712$\pm$0.024 & 0.607$\pm$0.021 \\
\multicolumn{2}{c}{}                   & Camera2 & 0.495$\pm$0.021 & 0.389$\pm$0.022 &    0.703$\pm$0.022 & 0.553$\pm$0.020 &   0.719$\pm$0.022 & 0.615$\pm$0.022 \\
\multicolumn{2}{c}{}                   & Camera3 & 0.412$\pm$0.025 & 0.377$\pm$0.023 &    0.532$\pm$0.023 & 0.512$\pm$0.022 &   0.672$\pm$0.026 & 0.572$\pm$0.023 \\
\multicolumn{2}{c}{}                  & All $3$ Cameras & 0.513$\pm$0.022 & 0.438$\pm$0.024 &    0.718$\pm$0.019 & 0.579$\pm$0.018 &   0.731$\pm$0.021 & 0.706$\pm$0.022 \\ \hline
%-------------------------------------------------------------------------------------------------------------------------------------------------------------------
\multicolumn{2}{c}{\multirow{4}{*}{P2}}& Camera1 & 0.422$\pm$0.023 & 0.387$\pm$0.020 &    0.604$\pm$0.024 & 0.526$\pm$0.021 &   0.622$\pm$0.022 & 0.518$\pm$0.020 \\
\multicolumn{2}{c}{}                   & Camera2 & 0.452$\pm$0.022 & 0.416$\pm$0.023 &    0.631$\pm$0.025 & 0.548$\pm$0.020 &   0.652$\pm$0.021 & 0.546$\pm$0.021 \\
\multicolumn{2}{c}{}                   & Camera3 & 0.378$\pm$0.021 & 0.351$\pm$0.022 &    0.517$\pm$0.022 & 0.435$\pm$0.023 &   0.538$\pm$0.025 & 0.441$\pm$0.022 \\
\multicolumn{2}{c}{}                  & All $3$ Cameras & 0.471$\pm$0.020 & 0.423$\pm$0.021 &    0.651$\pm$0.020 & 0.547$\pm$0.019 &   0.672$\pm$0.018 & 0.573$\pm$0.023 \\ \hline
%-------------------------------------------------------------------------------------------------------------------------------------------------------------------
\multicolumn{2}{c}{\multirow{1}{*}{P1\&P2}} & All $3$ Cameras & 0.524$\pm$0.032 & 0.475$\pm$0.031 &    0.802$\pm$0.028 & 0.651$\pm$0.025 &   0.892$\pm$0.019 & 0.751$\pm$0.020 \\ \hline
\end{tabular}
%}
\end{adjustbox}
\end{table}

%***********************************
\begin{table}[h!]
\caption{\small Average accuracy of FR systems using the proposed S+V model, SRC, and ESRC over different sessions and portals of the COX-S2V dataset. Feature representations are raw pixels, the 3DMM method is used for face synthesis.}
\label{table2}
\centering
\begin{adjustbox}{width=1\textwidth}
\begin{tabular}{c|ccccccc}
\hline
\multirow{3}{*}{\textbf{Viewpoint}} & \multicolumn{5}{c}{\textbf{Accuracy}} \\ %\cline{2-3} 
     &   \multicolumn{2}{c}{\textbf{SRC}}  &  \multicolumn{2}{c}{\textbf{ESRC}} & \multicolumn{2}{c}{\textbf{S+V Model}} & \\ \cline{2-7} 
                &    pAUC(20\%)   &       AUPR      &    pAUC(20\%)   &       AUPR      &    pAUC(20\%)   &    AUPR  \\ \hline \hline        
Camera1         & 0.481$\pm$0.020 & 0.432$\pm$0.021 & 0.765$\pm$0.019 & 0.645$\pm$0.022 & 0.780$\pm$0.020 & 0.657$\pm$0.021 \\
Camera2         & 0.475$\pm$0.023 & 0.419$\pm$0.022 & 0.716$\pm$0.020 & 0.602$\pm$0.020 & 0.747$\pm$0.023 & 0.629$\pm$0.022 \\
Camera3         & 0.507$\pm$0.021 & 0.441$\pm$0.019 & 0.802$\pm$0.021 & 0.671$\pm$0.021 & 0.824$\pm$0.021 & 0.715$\pm$0.019 \\ \hline
All 3 Cameras   & 0.566$\pm$0.030 & 0.480$\pm$0.027 & 0.835$\pm$0.027 & 0.695$\pm$0.026 & 0.905$\pm$0.020 & 0.776$\pm$0.017 \\ \hline
\end{tabular}
%\vspace{1cm}
\end{adjustbox}
\end{table}

%*****************************************************
\subsection{Impact of Feature Representations:}

Table~\ref{table3} shows the effect on FR performance of using different feature representations (including raw pixels, AlexNet~\cite{krizhevsky}, ResNet~\cite{He} and VGGNet~\cite{simonyan}) and face synthesis methods (3DMM and 3DMM-CNN) for videos from all $3$ cameras of the Chokepoint and COX-S2V datasets. 

\begin{table}[h!]
\caption{\small Average accuracy of FR systems using the proposed S+V model and template matching using different feature representation on Chokepoint and COX-S2V databases.}
\label{table3}
  \footnotesize
  \centering
  \begin{tabular}{l|l|l||cccc}
	\hline
	\multirow{3}{*}{\textbf{Technique}} & \multirow{3}{*}{\textbf{Face Synthesis}} & \multirow{3}{*}{\textbf{Features}} & \multicolumn{4}{c}{\textbf{Accuracy}}  \\  
    &   & & \multicolumn{2}{c}{\textbf{Chokepoint database}} & \multicolumn{2}{c}{\textbf{COX-S2V database}} \\ \cline{4-7}  
    &   &  & pAUC(20\%)    &       AUPR      &    pAUC(20\%)  &    AUPR  \\ \hline \hline
\multirow{5}{*}{TM} &\multirow{5}{*}{N/A}&  Raw pixels   & 0.551$\pm$0.027 & 0.503$\pm$0.028 & 0.574$\pm$0.031 & 0.512$\pm$0.029 \\ 
     &&    AlexNet  & 0.563$\pm$0.026 & 0.513$\pm$0.029 & 0.586$\pm$0.030 & 0.519$\pm$0.027 \\ 
     &&  VGGNet-16 & 0.570$\pm$0.028 & 0.524$\pm$0.026 & 0.597$\pm$0.027 & 0.528$\pm$0.030 \\  
     &&  VGGNet-19 & 0.578$\pm$0.025 & 0.531$\pm$0.027 & 0.605$\pm$0.029 & 0.533$\pm$0.028 \\ 
     &&   ResNet-50 & 0.595$\pm$0.027 & 0.550$\pm$0.026 & 0.628$\pm$0.024 & 0.551$\pm$0.025 \\ \hline
%-------------------------------------------------------------------------------------------------------
\multirow{5}{*}{SRC} &\multirow{5}{*}{N/A}&  Raw pixels & 0.525$\pm$0.030 & 0.475$\pm$0.029 & 0.568$\pm$0.031 & 0.481$\pm$0.030 \\ 
     &&  AlexNet   & 0.537$\pm$0.025 & 0.487$\pm$0.028 & 0.581$\pm$0.027 & 0.494$\pm$0.026 \\ 
     &&  VGGNet-16& 0.552$\pm$0.026 & 0.491$\pm$0.027 & 0.590$\pm$0.025 & 0.505$\pm$0.027 \\  
     &&  VGGNet-19& 0.567$\pm$0.027 & 0.512$\pm$0.024 & 0.602$\pm$0.023 & 0.511$\pm$0.028 \\ 
     &&  ResNet-50 & 0.581$\pm$0.026 & 0.533$\pm$0.025 & 0.623$\pm$0.022 & 0.523$\pm$0.024 \\ \hline
% ------------------------------------------------------------------------------------------------------          
    &\multirow{5}{*}{3DMM} &  Raw pixels & 0.892$\pm$0.018 & 0.751$\pm$0.019 & 0.903$\pm$0.020 & 0.775$\pm$0.016 \\
\multirow{8}{*}{S+V Model} &  & AlexNet  & 0.905$\pm$0.019 & 0.771$\pm$0.020 & 0.913$\pm$0.016 & 0.783$\pm$0.015 \\ 
     & & VGGNet-16 & 0.908$\pm$0.016 & 0.773$\pm$0.017 & 0.916$\pm$0.018 & 0.786$\pm$0.016 \\
     & & VGGNet-19 & 0.912$\pm$0.017 & 0.779$\pm$0.018 & 0.921$\pm$0.016 & 0.791$\pm$0.017 \\ 
     & & ResNet-50  & \textbf{0.917$\pm$0.015} & \textbf{0.783$\pm$0.016} & \textbf{0.925$\pm$0.015} & \textbf{0.798$\pm$0.014} \\ \cline{2-7} 
     & \multirow{5}{*}{3DMM-CNN} &  Raw pixels & 0.855$\pm$0.019 & 0.737$\pm$0.018 & 0.871$\pm$0.019 & 0.741$\pm$0.018 \\ 
     &&    AlexNet  & 0.873$\pm$0.020 & 0.752$\pm$0.020 & 0.884$\pm$0.018 & 0.753$\pm$0.019 \\ 
     && VGGNet-16& 0.880$\pm$0.017 & 0.759$\pm$0.017 & 0.891$\pm$0.017 & 0.761$\pm$0.016 \\  
     &&    VGGNet-19   & 0.884$\pm$0.018 & 0.763$\pm$0.020 & 0.902$\pm$0.016 & 0.765$\pm$0.017 \\ 
     &&   ResNet-50 & 0.891$\pm$0.016 & 0.769$\pm$0.014 & 0.907$\pm$0.017 & 0.771$\pm$0.015 \\ \hline
	\end{tabular}
\end{table}

%*******************************************
%*******************************************
%*******************************************
\vspace{2cm}
We further evaluate the impact on the performance of different CNN feature extractors and loss functions for FR with the S+V model. Table 4 shows the average AUC and AUPR accuracy of FR systems using the proposed S+V model with different pre-trained CNNs for feature representation and loss functions (triplet loss \cite{facenet}, cosine loss \cite{a} and angular softmax \cite{b}) on the Chokepoint and COX-S2V databases. Results indicate that coupling the S+V model with deep CNN features can further improve FR accuracy over using raw pixels, and that using ResNet-50 outperforms there other CNN  architectures. Additionally, SphereFace training method yields the higher accuracy. By using CNN features along with 3DMM or 3DMM-CNN, a still-to-video FR system with the S+V model outperforms the baseline template matcher (TM) and SRC.

Results show that coupling the S+V model with deep CNN features can further improve the FR accuracy over using raw pixels, and that using ResNet-50 outperforms all other deep architectures. The results also indicate that SphereFace training method yields higher accuracy. Using CNN features and 3DMM or 3DMM-CNN, a FR system with the S+V model outperform the baseline template matcher (TM) and SRC.

\begin{table}[h!]
\caption{Average accuracy of FR systems using the proposed S+V model (3DMM face synthesis) with different deep feature representations on Chokepoint and COX-S2V databases.}
\label{table31}
   \footnotesize
   \centering
   \begin{adjustbox}{width=1.1\textwidth}
   \begin{tabular}{l|l|l||cccc}
   \hline
   \multirow{3}{*}{\textbf{Technique}} & \multirow{3}{*}{\textbf{Deep Architecture}} & \multirow{3}{*}{\textbf{Training}} & \multicolumn{4}{c}{\textbf{Accuracy}}  \\  
     &  & & \multicolumn{2}{c}{\textbf{Chokepoint database}} & \multicolumn{2}{c}{\textbf{COX-S2V database}} \\ \cline{4-7}  
                              &  &  &    pAUC(20\%)    &     AUPR     &    pAUC(20\%)  &    AUPR  \\ \hline \hline
 & \multirow{3}{*}{AlexNet}   &    FaceNet \cite{facenet}  & 0.905$\pm$0.019 & 0.771$\pm$0.020 & 0.913$\pm$0.016 & 0.783$\pm$0.015 \\
   \multirow{7}{*}{S+V Model} & &   CosFace \cite{b}  & 0.908$\pm$0.021 & 0.774$\pm$0.022 & 0.915$\pm$0.017 & 0.787$\pm$0.016 \\ 
                              & & SphereFace \cite{a} & 0.912$\pm$0.020 & 0.780$\pm$0.018 & 0.918$\pm$0.015 & 0.792$\pm$0.014 \\ \cline{2-7} 
                              %------------------------------------------------------------------------------------------
     &\multirow{3}{*}{VGGNet-19} &    FaceNet \cite{facenet}  & 0.884$\pm$0.021 & 0.763$\pm$0.020 & 0.902$\pm$0.019 & 0.765$\pm$0.018 \\
                              & &   CosFace \cite{b}  & 0.889$\pm$0.019 & 0.768$\pm$0.022 & 0.907$\pm$0.017 & 0.772$\pm$0.016 \\ 
                              & & SphereFace \cite{a} & 0.906$\pm$0.018 & 0.771$\pm$0.017 & 0.913$\pm$0.015 & 0.778$\pm$0.017 \\ \cline{2-7}
                              %------------------------------------------------------------------------------------------
 & \multirow{3}{*}{ResNet-50} &    FaceNet \cite{facenet}  & 0.917$\pm$0.015 & 0.783$\pm$0.016 & 0.924$\pm$0.015 & 0.798$\pm$0.014 \\ 
                              & &   CosFace \cite{b} & 0.920$\pm$0.018 & 0.786$\pm$0.019 & 0.927$\pm$0.018 & 0.802$\pm$0.020 \\ 
                              & & SphereFace \cite{a} & 0.922$\pm$0.015 & 0.791$\pm$0.014 & 0.928$\pm$0.017 & 0.805$\pm$0.015 \\ \hline
	\end{tabular}
	\end{adjustbox}
\end{table}
%*******************************************

Tables \ref{table51} shows the average accuracy of FR for the augmented and auxiliary dictionaries with the videos from all $3$ cameras of the Chokepoint and COX-S2V datasets, respectively.
\begin{table}[h!]
\caption{Average accuracy of FR systems using the augmented dictionary (3DMM face synthesis) and auxiliary dictionaries on Chokepoint and COX-S2V databases.}
\label{table51}

  \footnotesize
  \centering
  \begin{tabular}{ll|lcccc}
	\hline
	& \multirow{3}{*}{\textbf{Technique}}   & \multicolumn{4}{c}{\textbf{Accuracy}}  \\  
    & & \multicolumn{2}{c}{\textbf{Chokepoint database}} & \multicolumn{2}{c}{\textbf{COX-S2V database}} \\ \cline{3-6}  
                    &  & pAUC(20\%)    &     AUPR     &    pAUC(20\%)    &    AUPR  \\ \hline \hline
\multirow{2}{*}{S+V Model} & Augmented Dictionary & 0.829$\pm$0.28 & 0.705$\pm$0.27 & 0.847$\pm$0.26 & 0.718$\pm$0.254 \\ 
                           & Auxiliary Dictionary & 0.836$\pm$0.23 & 0.714$\pm$0.25 & 0.862$\pm$0.22 & 0.731$\pm$0.021 \\ \hline
	\end{tabular}
\end{table}

%*******************************************
\subsection{Comparison with State-of-the-Art Methods:}
 
Table~\ref{table4} presents the FR accuracy obtained with the proposed S+V model compared with the state-of-the-art SRC techniques based on generic learning -- {ESRC}~\cite{Deng}, {SVDL}~\cite{Yang}, {LGR}~\cite{Zhu}, {RADL}~\cite{Wei}, {CSR}~\cite{Li}. Each one uses the same number of samples, raw pixel-based features, and a regularization parameter $\Lambda$ set to $0.005$. Accuracy of the S+V model is also compared with that of the Flow-Based Face Frontalization~\cite{Hassner} and Recognition via Generation~\cite{masi2} techniques. The baseline system is a SRC model designed with the original reference still ROI of each enrolled person, and raw pixel-based features. The table shows that the S+V model, using a joint generic learning and face synthesis, achieves the higher level of accuracy than other methods under the same configuration, has potential in surveillance FR. 

\begin{table}[t!]
   \vspace{.5cm}
	\caption{\small Average accuracy of FR systems based on the proposed S+V model and related state-of-the art SRC methods for videos from all 3 cameras of the Chokepoint and COX-S2V databases. Feature representations are raw pixels, the 3DMM method is used for face synthesis.}
	\label{table4}
    \footnotesize
	\centering
		\begin{tabular}{l||ccccc|}
			 \hline
			 \multirow{3}{*}{\textbf{Techniques}}  & \multicolumn{4}{c}{\textbf{Accuracy}}  \\ %\cline{3-6} 
              &     \multicolumn{2}{c}{\textbf{Chokepoint database}}   & \multicolumn{2}{c}{\textbf{COX-S2V database}}   \\ \cline{2-5}      
			                        &   pAUC(20\%)    &    AUPR   &    pAUC(20\%)   &    AUPR \\ \hline \hline
			  SRC (Baseline) {\cite{Wright1}}  & 0.524$\pm$0.032 & 0.475$\pm$0.031  &  0.568$\pm$0.030 & 0.480$\pm$0.027 \\
			  ESRC {\cite{Deng}}    & 0.802$\pm$0.028 & 0.651$\pm$0.025  &  0.835$\pm$0.027 & 0.695$\pm$0.026 \\
			  ESRC-KSVD             & 0.811$\pm$0.023 & 0.659$\pm$0.022  &  0.840$\pm$0.023 & 0.712$\pm$0.021 \\
			  SVDL {\cite{Yang}}    & 0.825$\pm$0.023 & 0.703$\pm$0.025  &  0.843$\pm$0.025 & 0.724$\pm$0.023 \\
			  RADL {\cite{Wei}}     & 0.832$\pm$0.019 & 0.711$\pm$0.020  &  0.849$\pm$0.022 & 0.730$\pm$0.021 \\
			  LGR {\cite{Zhu}}      & 0.849$\pm$0.022 & 0.717$\pm$0.024  &  0.878$\pm$0.023 & 0.744$\pm$0.025 \\
			  CSR {\cite{Li}}       & 0.852$\pm$0.025 & 0.722$\pm$0.020  &  0.880$\pm$0.021 & 0.753$\pm$0.020 \\ 
Face Frontalization {\cite{Hassner}}& 0.822$\pm$0.021 & 0.711$\pm$0.023  &  0.843$\pm$0.022 & 0.719$\pm$0.023 \\ 
Recognition via Generation {\cite{masi2}}& 0.815$\pm$0.023 & 0.703$\pm$0.025  &  0.838$\pm$0.024 & 0.705$\pm$0.026 \\
	\textbf{S+V Model (Ours)}    & \textbf{0.892$\pm$0.019} & \textbf{0.751$\pm$0.020}  &  \textbf{0.905$\pm$0.018} & \textbf{0.776$\pm$0.017} \\ 
			 %S+V Model    &  ResNet-50 & \textbf{0.917$\pm$0.018} & \textbf{0.783$\pm$0.019} & \textbf{0.925$\pm$0.018} & \textbf{0.798$\pm$0.016 
			 \\ \hline
		\end{tabular}
\end{table}

%*****************************************************
%\subsubsection{Face Recognition under Different Poses:}

\begin{figure}[!b]
\vspace{-.1cm}
	\centering 
	\includegraphics[width=\textwidth]{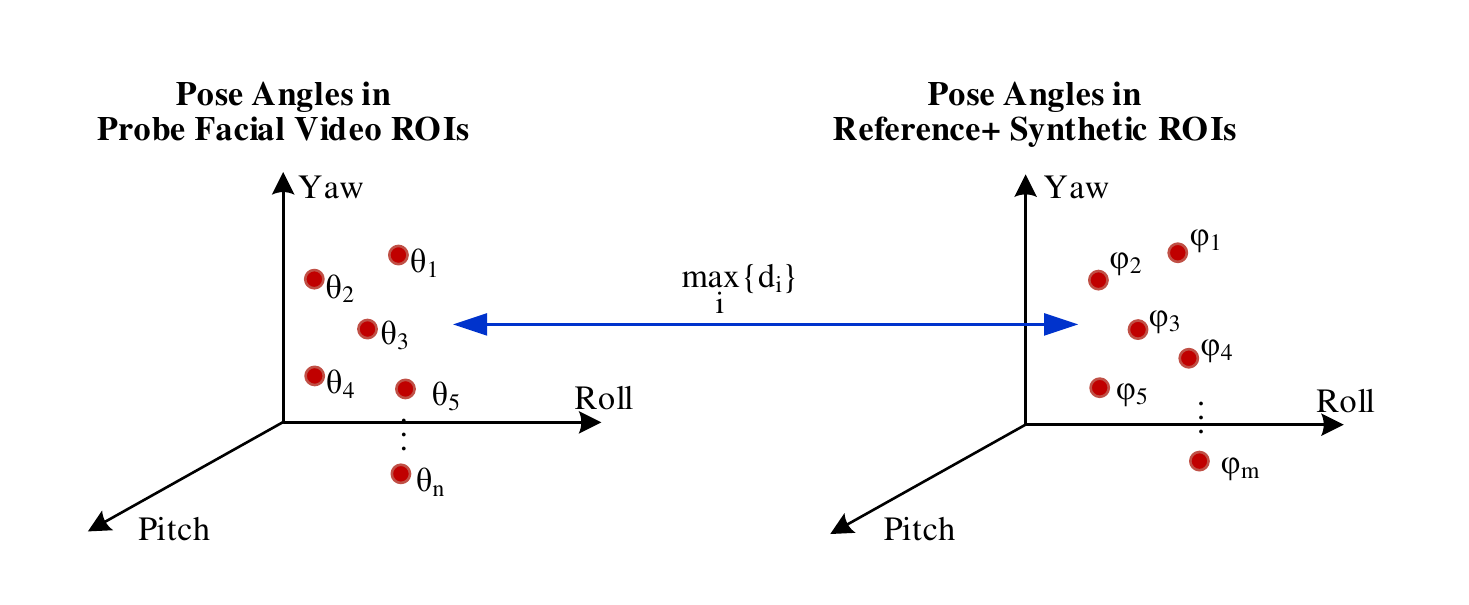}
	\vspace{-.4cm}
	\caption{\small Illustration of procedure for the selection of the largest pose variations.} 
    \vspace{.7cm}
	\label{fig12}
\end{figure} 

%-----------------------
\begin{figure}[!b]
    \vspace{-.2cm}
   \centering
   \subfigure[Chokepoint]{\label{fig:a}\includegraphics[width=58mm]{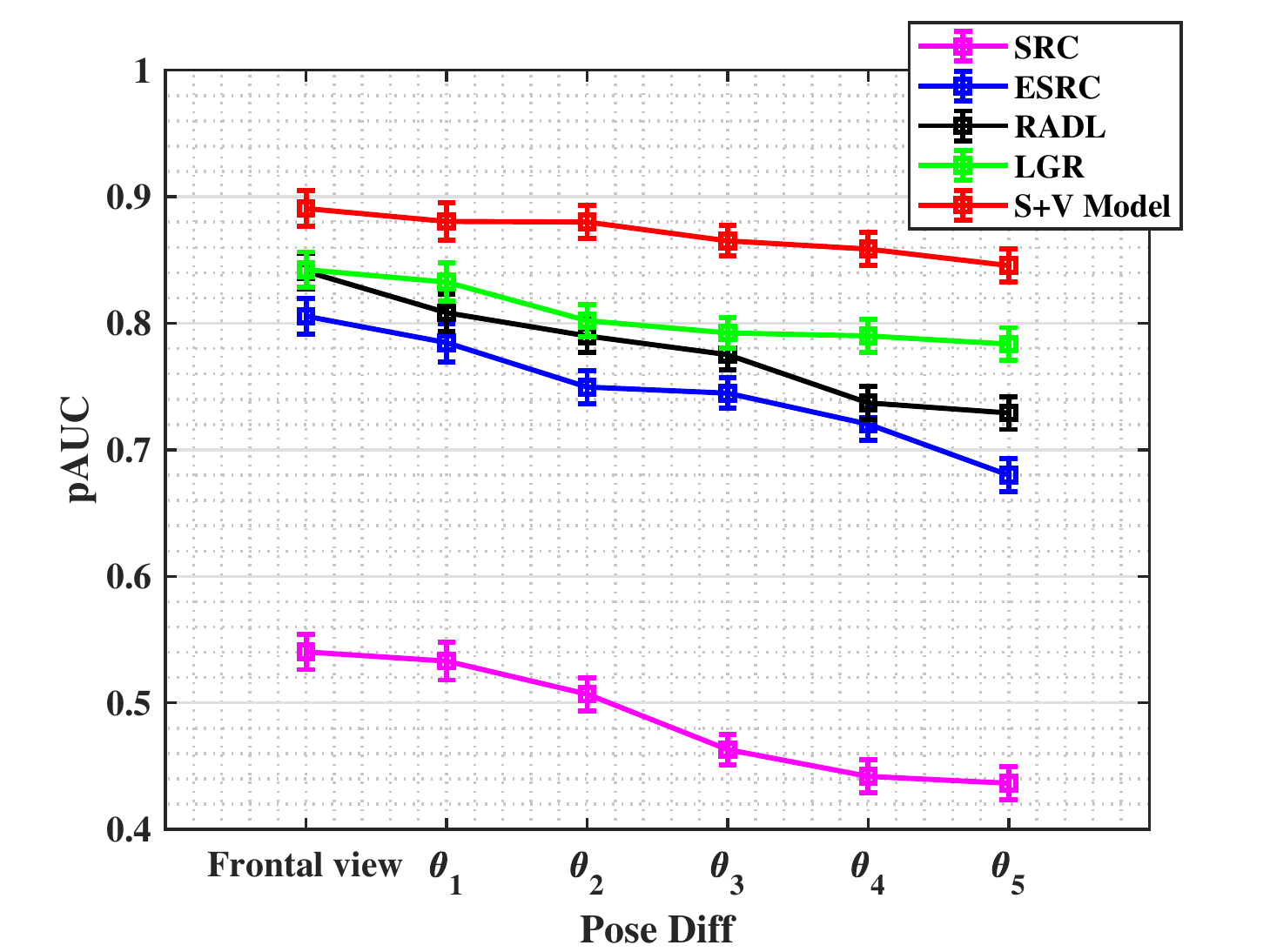}}
   \vspace{-.1cm}
   ~ 
   \centering
   \subfigure[Chokepoint]{\label{fig:a}\includegraphics[width=58mm]{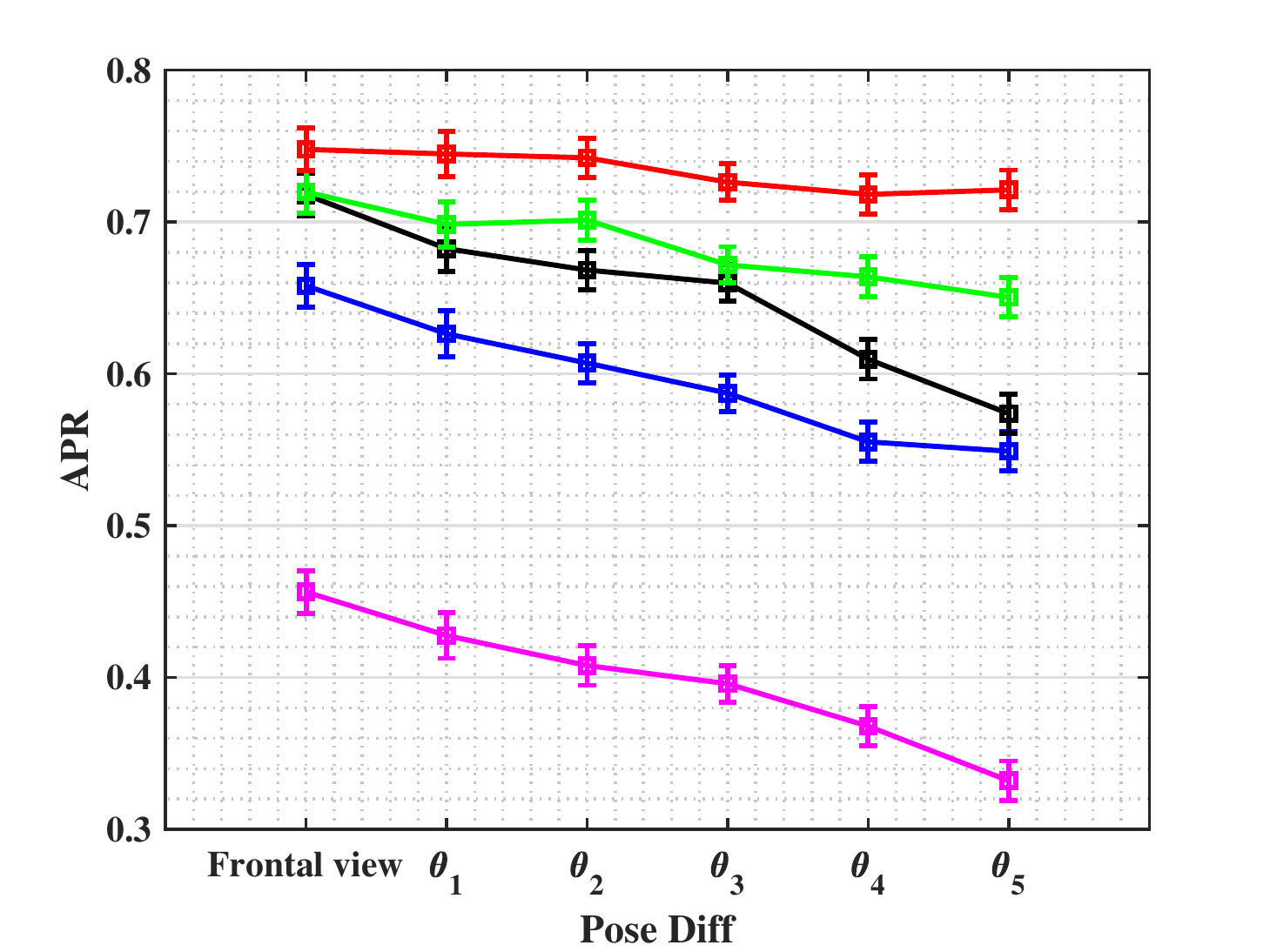}}
   \vspace{-.1cm}
    ~
   \centering
   \subfigure[COX-S2V]{\label{fig:a}\includegraphics[width=58mm]{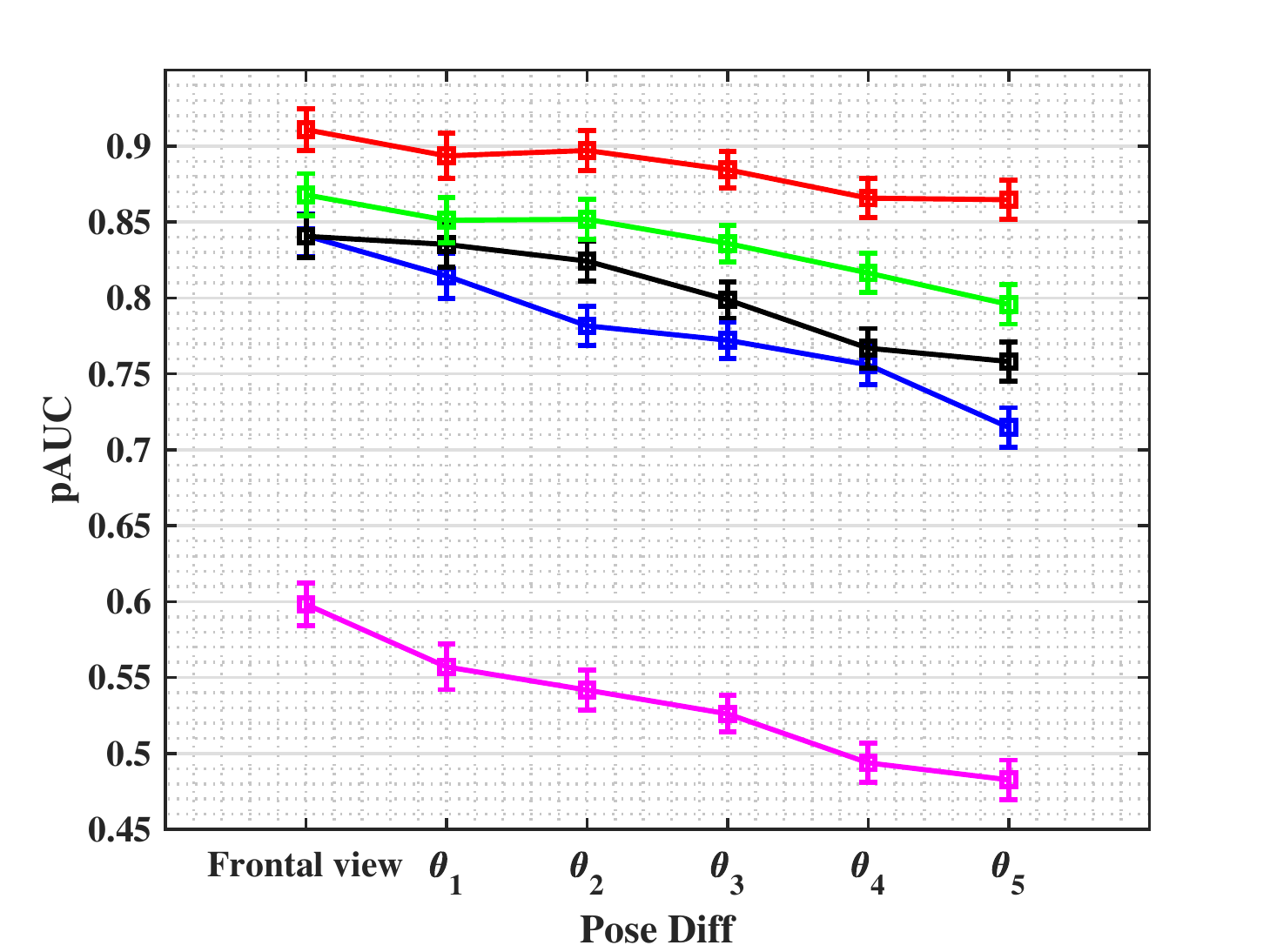}}
   ~ 
   \centering
   \subfigure[COX-S2V]{\label{fig:a}\includegraphics[width=58mm]{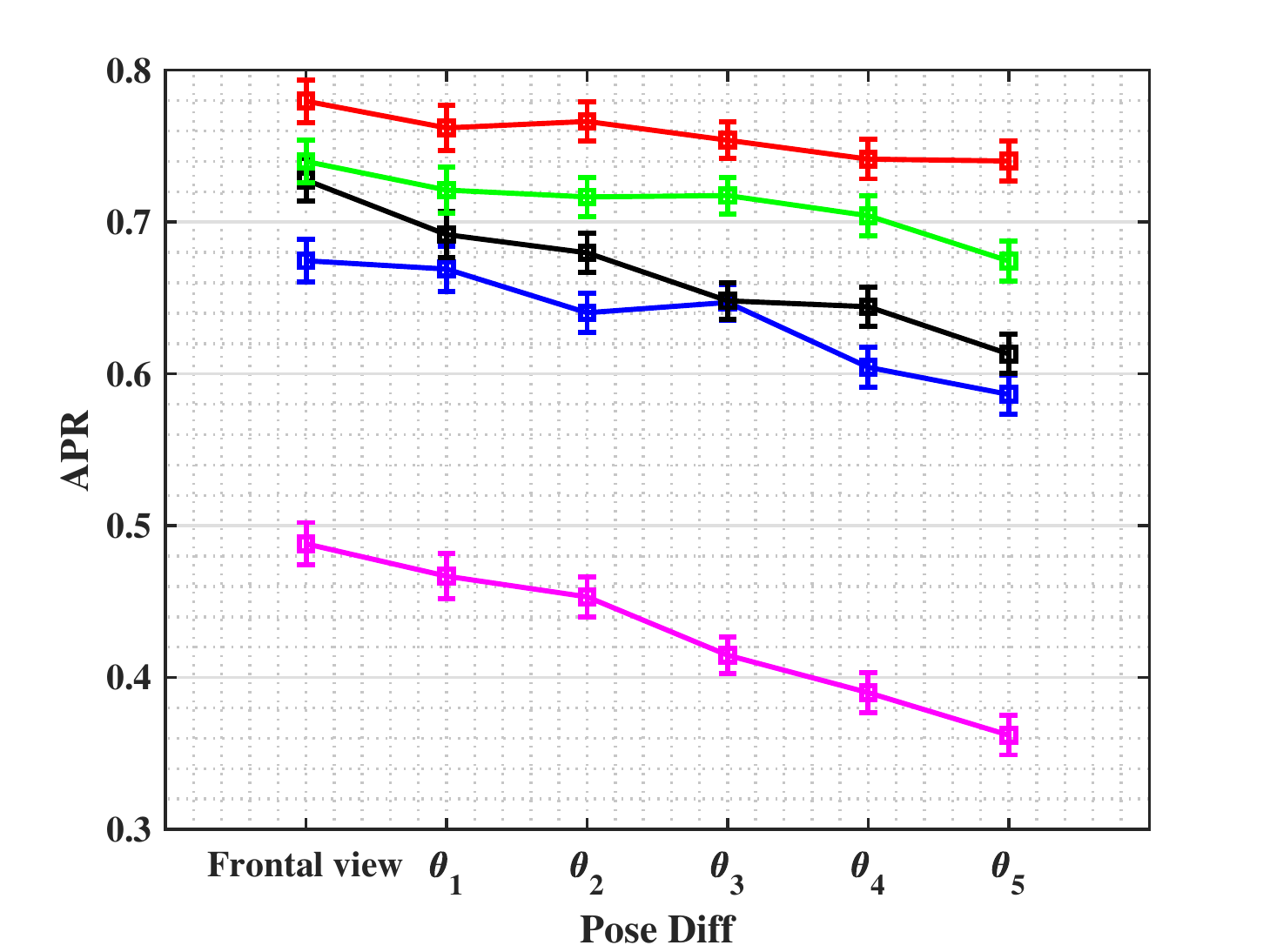}}
   \caption{\small Average pAUC(20\%) and AUPR accuracy of S+V model and related state-of-the-art techniques versus the different pose variations on Chokepoint (a,b) and COX-S2V (c,d) databases. Error bars are standard deviation.} 
   \label{fig13} 
\end{figure}

In order to assess still-to-video FR accuracy under the worst-case pose variations between the probe video ROIs and augmented gallery dictionary ROIs, we compute the minimum distance between the pose angle of each probe video ROI ($20$ trajectories in $3$ cameras), $\{ \theta_1, \theta_2, \dots , \theta_n \}$, and pose angles of both reference still  and synthetic ROIs in the augmented gallery dictionary, $\{\varphi_1, \varphi_2, \dots , \varphi_m \}$:
\begin{equation}
d_i = \min_{j} \{ {\parallel (\theta_i - \varphi_j) \parallel} : j = 1, 2, \dots, m \} , 
\end{equation}
where $d_i$ corresponds to the $i^{th}$ probe video ROI, for $i= 1, 2, \dots, n$. Next, 5 video ROIs that have the largest distance, $\max \limits_{i} \{ d_i \}$, are chosen as the faces with the largest pose differences (see Fig.~\ref{fig12}). 
%
%\begin{equation}
%R_d^i = \max_{i}( \min_{j} \{ {\parallel (\theta_i - \varphi_j) \parallel}) \quad :  \quad j = 1, \dots, m \} .
%\end{equation}
%
Fig.~\ref{fig13} shows the accuracy obtained with the SRC, ESRC, RADL, LGR and S+V models when these ROIs are classified as probe ROIs.

As the pose differences increase, FR accuracy decreases. The FR system using the S+V model reaches the highest accuracy due to the added robustness to pose variations. Then, LGR outperforms SRC, ESRC and RADL across all pose variations. Accuracy of the SRC is much lower than the others because, with only one frontal reference gallery ROI per person, the probe ROIs are not well represented.

\begin{figure}[ht]
   \centering
   \subfigure[Chokepoint]{\label{fig:a}\includegraphics[width=58mm]{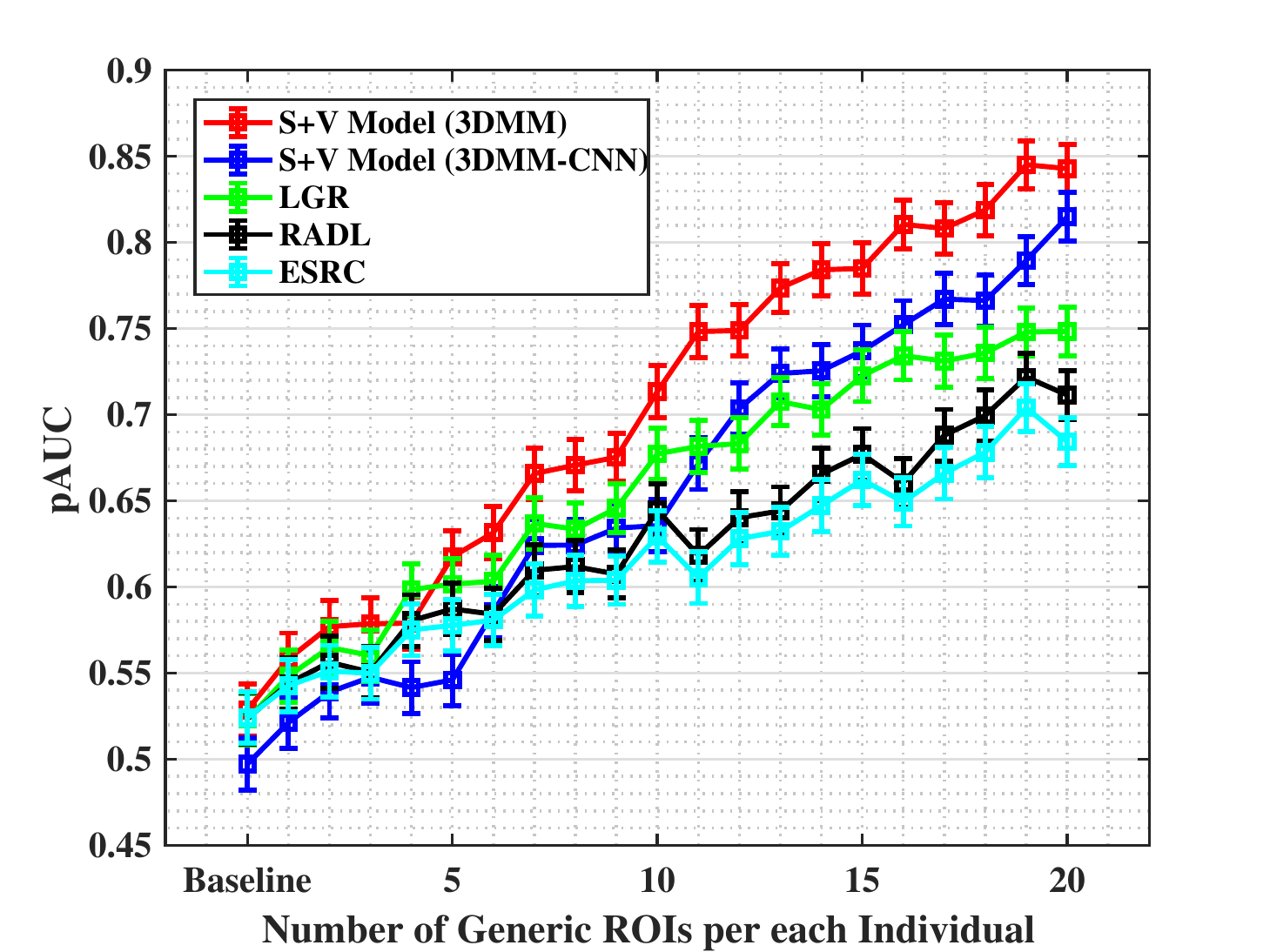}}
   ~ 
   \centering
   \subfigure[Chokepoint]{\label{fig:a}\includegraphics[width=58mm]{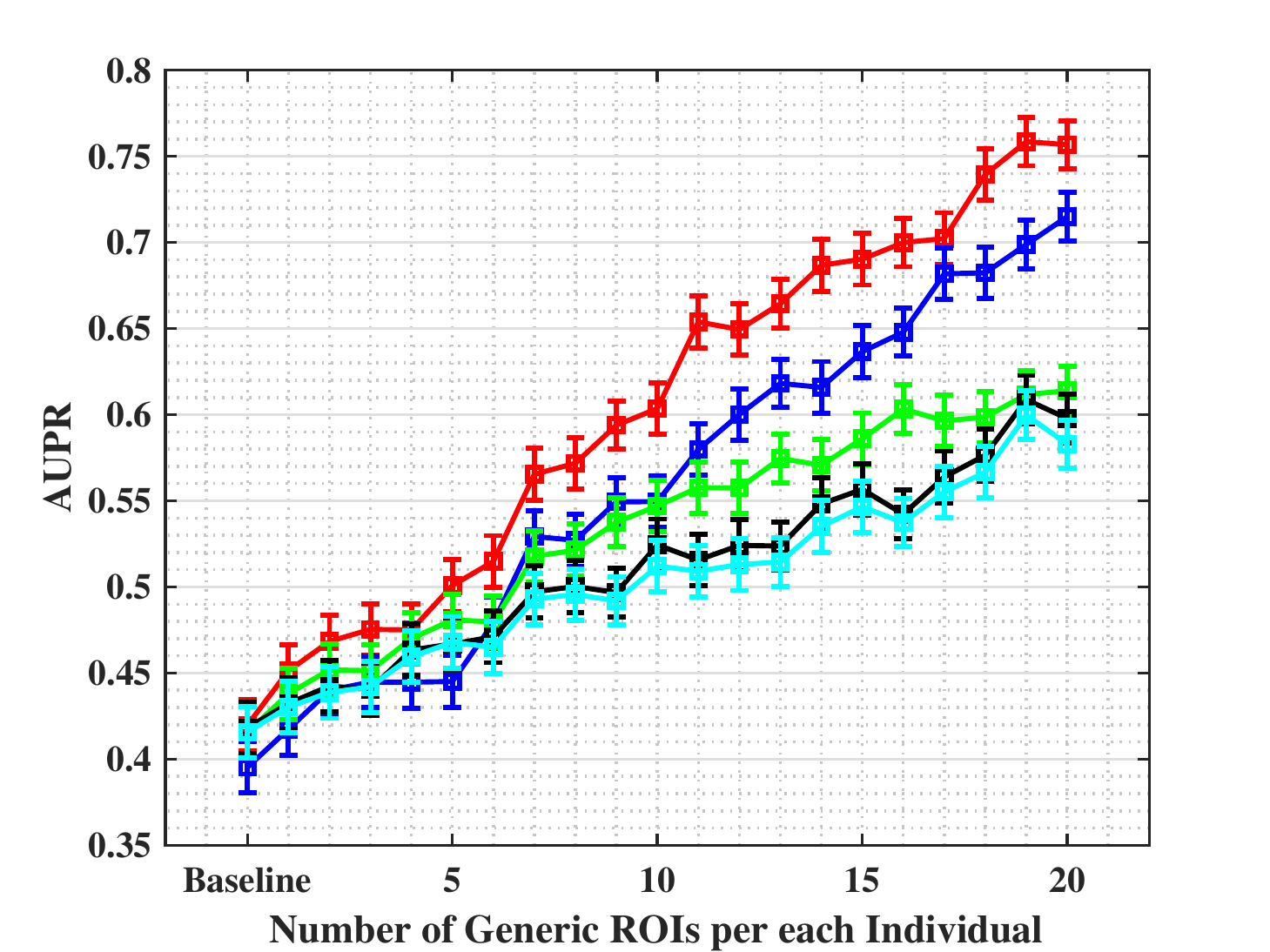}}
    ~
   \centering
   \subfigure[COX-S2V]{\label{fig:a}\includegraphics[width=58mm]{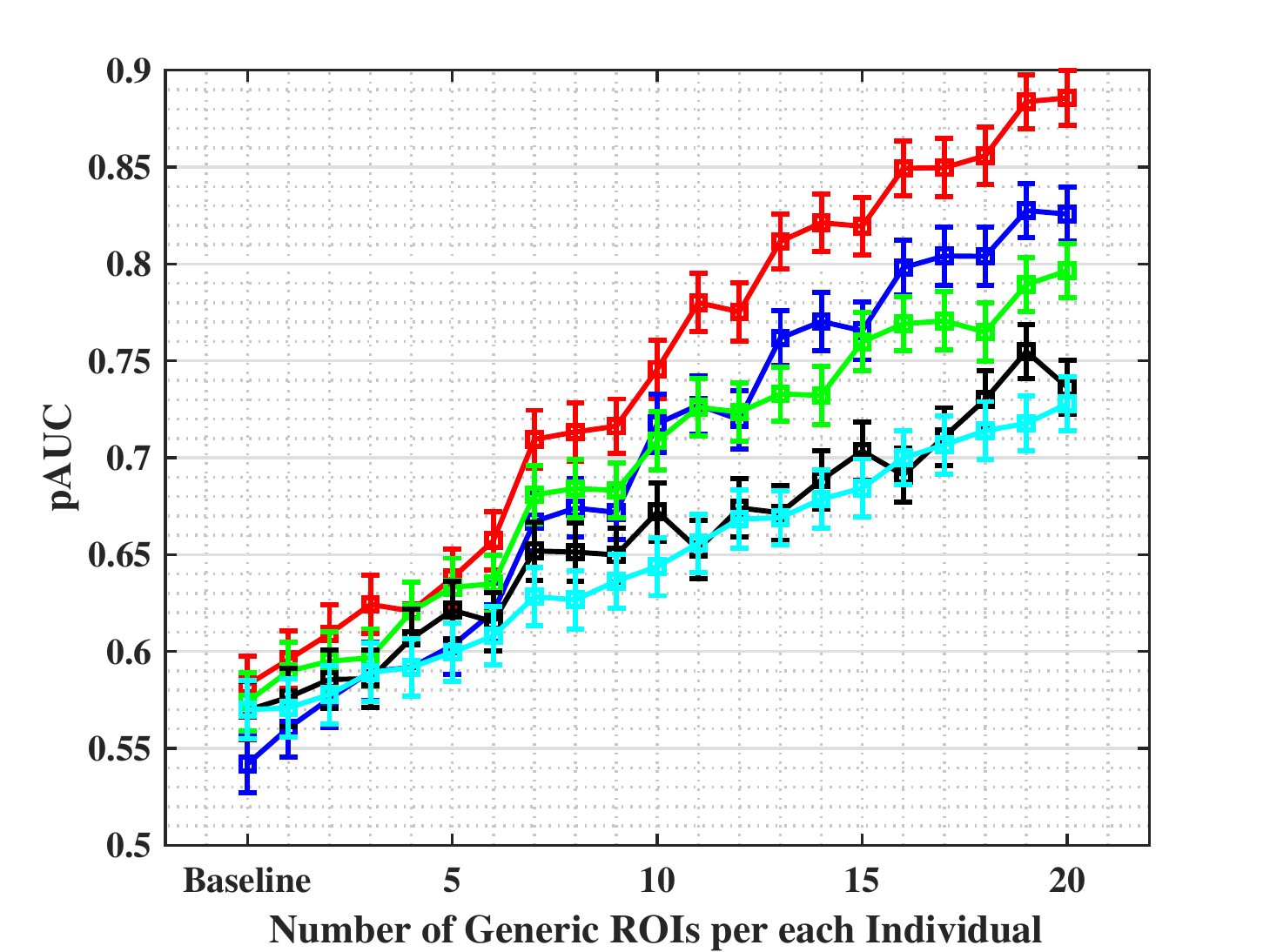}}
   ~ 
   \centering
   \subfigure[COX-S2V]{\label{fig:a}\includegraphics[width=58mm]{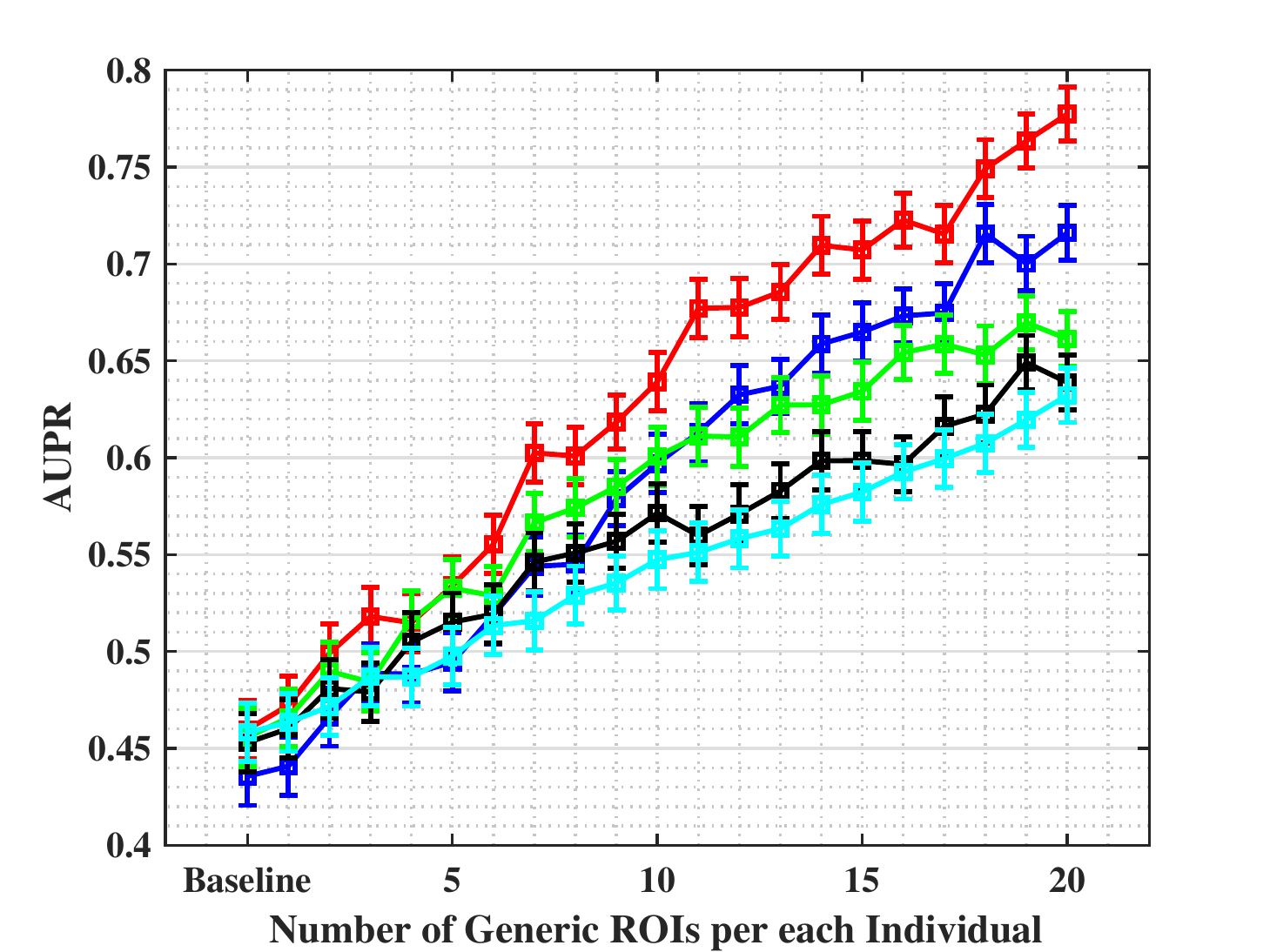}}
   \caption{\small Average pAUC(20\%) and AUPR accuracy versus the size of the generic set on Chokepoint (a,b) and COX-S2V (c,d)    databases. Error bars are standard deviation.} 
   \label{fig14} 
\end{figure}

%************************************************************************
%\subsubsection{Performance based on the Size of Generic Set:}
Fig.~\ref{fig14} shows the impact of the size of generic set in the auxiliary variational dictionary on FR accuracy. The results of SRC, ESRC, RADL and LGR are also shown for the same configurations for comparison. Accuracy of the S+V model increases significantly with respect to other state-of-the-art methods as the number of generic ROIs grows. The results support the conclusion that by augmenting the gallery dictionary, allows the S+V model to increasingly benefit from the variational information of the generic set.

%*******************************************************************
\subsection{Ablation Study:}

Designing S+V model for still-to-video FR consists of three main steps: ($\mathcal{M}_1$) face synthesis, ($\mathcal{M}_2$) adding intra-class variations, and ($\mathcal{M}_3$) pairing the dictionaries. In this subsection, an ablation study is presented to show the impact of each module on the FR performance. We assume that all FR systems use a pixel-based feature representation, 3DMM face synthesis, and $q$ synthetic images in the augmented dictionary.

Tables~\ref{table5} and~\ref{table6} shows the average accuracy of the ablation study with videos from all $3$ cameras of the Chokepoint and COX-S2V datasets, respectively. Firstly, we disabled the face synthesis module, $\mathcal{M}_1$, and performed experiments to show the impact of augmenting the reference gallery with synthetic faces on FR accuracy. Next, we removed the auxiliary dictionary to evaluate the impact of considering generic set variations with the S+V model. By removing both $\mathcal{M}_1$ and $\mathcal{M}_2$ modules from the S+V model, accuracy declines significantly by about $50\%$. The results suggest that the addition of synthetic and generic set faces is an effective strategy to cope with facial variations. Another important component of the S+V model is the selection of representative ROIs and pairing the dictionaries. By removing the row sparsity and joint sparsity in the S+V model, $\mathcal{M}_3$, and by adding $10$ randomly selected synthetic ROIs, accuracy decreases by about $15\%$.  

\begin{table}[h!]
\caption{\small The results of ablation study with Chokepoint database.}
\label{table5}
\footnotesize
\centering
\begin{tabular}{c||cccc}
\hline
\bf{Accuracy}       & \multicolumn{4}{c}{\bf{Removed Module}}    \\ 
                    & baseline (none)   & $\mathcal{M}_1$ & $\mathcal{M}_2$ & $\mathcal{M}_3$   \\ \hline \hline
pAUC(20\%)          & 0.892$\pm$0.019   &  0.839$\pm$0.21 & 0.827$\pm$0.27  & 0.883$\pm$0.25    \\ \hline
AUPR                & 0.751$\pm$0.020   &  0.709$\pm$0.23 & 0.702$\pm$0.25  & 0.721$\pm$0.22    \\ \hline

\end{tabular}
\end{table}

%----------------------------------------------------------------------------
\begin{table}[h!]
\caption{\small The results of ablation study with COX-S2V database.}
\label{table6}
\footnotesize
\centering
\begin{tabular}{c||cccc}
\hline
\bf{Accuracy}       & \multicolumn{4}{c}{\bf{Removed Module}}    \\ 
                    & baseline (none)   & $\mathcal{M}_1$   & $\mathcal{M}_2$   & $\mathcal{M}_3$   \\ \hline \hline
pAUC(20\%)          & 0.905$\pm$0.018   & 0.857$\pm$0.22    & 0.835$\pm$0.24    & 0.887$\pm$0.20 \\ \hline
AUPR                & 0.776$\pm$0.017   & 0.721$\pm$0.20    & 0.712$\pm$0.21    & 0.769$\pm$0.21 \\ \hline

\end{tabular}
\end{table}

% \begin{table}[h!]
% \caption{\small Quantitative results of ablation study with Chokepoint and COX-S2V databases.}
% \label{table5}
% \footnotesize
% \centering
% \begin{tabular}{lllllllll}
% \hline
%                  & \multicolumn{4}{c}{Chokepoint database} & \multicolumn{4}{c}{COX-S2V database} \\ \hline
% \multicolumn{1}{l|}{Removed Module}&          -      & $\mathcal{M}_1$ & $\mathcal{M}_2$ & $\mathcal{M}_3$ &         -       & $\mathcal{M}_1$ & $\mathcal{M}_2$ & $\mathcal{M}_3$ \\ \hline
% \multicolumn{1}{l|}{pAUC}          & 0.892$\pm$0.019 &  0.839$\pm$0.21 & 0.827$\pm$0.27  & 0.883$\pm$0.25  & 0.905$\pm$0.018 & 0.857$\pm$0.022 & 0.835$\pm$0.024 & 0.887$\pm$0.020 \\ \hline
% \multicolumn{1}{l|}{AUPR}          & 0.751$\pm$0.020 &  0.709$\pm$0.23 & 0.702$\pm$0.25  & 0.721$\pm$0.22  & 0.776$\pm$0.017 & 0.721$\pm$0.020 & 0.712$\pm$0.021 & 0.769$\pm$0.021 \\ \hline
% \end{tabular}
% \end{table}

%*********************************
\subsection{Complexity Analysis:}
Time complexity is an important consideration in many real-time FR applications in video surveillance. The time required by the S+V model to classify a probe ROI is $\mathcal{O} (d(N+M)Lq \log{} n + Lk(q+1))$  where $d$ is the dimension of the face descriptors, $n$ is the number of ROIs per class in the augmented gallery dictionary, $k$ is the total number of classes (enrolled individuals), $N = k n$ is the total number of reference still images, $M$ is the total size of the external generic set, $q$ is the number of views, and $L$ is number active sets (at each iteration, we need to select $L$ most representative dynamic active sets from coefficient matrix.) In video FR applications, $N$ may be larger, therefore the computational burden of handling larger dictionaries may represent bottleneck of the proposed method. 
The complexity of SRC and ESRC are $\mathcal{O} (d^2N)$, $\mathcal{O} (d^2(N+M))$, respectively. The complexity of LGR is $\mathcal{O} (s({n_d}^3+{n_d}^2{d_p}))$ where $s$ is the number of patches, $n_d$ is the total number of patches, $d_p$ is the feature dimension of patches. Although the proposed S+V model outperforms SRC and ESRC, it requires more computations, mostly because of the pairing of the dictionaries. 

Table~\ref{table7} reports the average test time required by the proposed and baseline techniques to classify a probe ROI from Chokepoint and COX-S2V videos. The LGR and RADL are more computationally intensive than the S+V model. Finally, Table~\ref{table8} reports the average time for the $3$ main steps of the proposed framework: face synthesis ($\mathcal{M}_1$), intra-class variation extraction ($\mathcal{M}_2$), and pairing the dictionaries ($\mathcal{M}_3$) on videos of all 3 cameras in the Chokepoint and COX-S2V datasets.
% Finally, Table~\ref{table8} reports the average time of different steps of the proposed framework individually on Chokepoint and COX-S2V.
The time complexity of $\mathcal{M}_1$ is the highest, followed by $\mathcal{M}_3$ with complexity $\mathcal{O}({M N log(M)})$, where $M$ and $N$ are, respectively, the number of rows and columns of the dissimilarity matrix. 

\begin{table}[ht]
\caption{\small Average time required by techniques to classify a probe videos ROI with the Chokepoint and COX-S2V datasets.}
\label{table7}
\footnotesize
\centering
\begin{tabular}{l|cc}
\hline
\multirow{2}{*}{\textbf{Technique}}  &   \multicolumn{2}{c}{\textbf{Classification Time (sec)}} \\ \cline{2-3} 
 			 		 & \textbf{Chokepoint database}	 &  \textbf{COX-S2V database} 	\\ \hline \hline   
SRC~\cite{Wright1}   &   1.03   &    2.56     \\
ESRC~\cite{Deng}     &   1.72   &    3.42     \\
RADL~\cite{Wei}      &   4.62   &    8.15     \\
LGR~\cite{Zhu}       &   7.13   &   12.37     \\ 
S+V Model            &   2.81   &    4.83     \\ 
\hline
\end{tabular}
\end{table}

%-------------------------------------
\begin{table}[ht]
\caption{\small Average computational time of different step in the S+V model with the Chokepoint and COX-S2V datasets.}
\label{table8}
\footnotesize
\centering
\begin{tabular}{l|cc}
\hline
\multirow{2}{*}{\textbf{Module}}  &  \multicolumn{2}{c}{\textbf{Processing Time (Sec)}} \\ \cline{2-3} 
      & \textbf{Chokepoint database}  &  \textbf{COX-S2V database} \\ \hline \hline   
$\mathcal{M}_1$ (3DMM)        &  120  &  120  \\
$\mathcal{M}_1$ (3DMM-CNN)    &  1.3  &  1.3  \\
$\mathcal{M}_2$               &  0.53 &  0.53 \\
$\mathcal{M}_3$               &  2.47 &  4.41  \\
\hline
\end{tabular}
\end{table}

%****************************************
\section{Conclusion}
\label{conclusion}
%****************************************

In this paper, a paired sparse reconstruction model is proposed to account for linear and non-linear variations in the context of still-to-video FR. The proposed S+V model leverages both face synthesis and generic learning to effectively represent probe ROIs from a single reference still. This approach manages the non-linear variations by enriching the gallery dictionary with a representative set of synthetic profile faces, where synthetic (still) faces are paired with generic set (video) face in the auxiliary variational dictionary. In this way, the augmented gallery dictionary is encouraged to share the same sparsity pattern with the auxiliary dictionary for the same pose angles.
%This paper proposes a method for matching still image and unconstrained video. A probe face is assumed to be the linear combination of faces in the gallery dictionary and faces in the variational dictionary. The gallery dictionary is augmented with synthetic faces under different poses. The coefficients with respect both dictionaries share the same sparsity.
%
Experimental results obtained using the Chokepoint and COX-S2V datasets suggest that the proposed S+V model allows us to efficiently represent linear and non-linear variations in facial pose with no need to collect a large amount of training data, and with only a moderate increase in time complexity. Results indicated that generic learning alone cannot effectively resolve the challenges of the SSPP and visual domain shift problems. With S+V model, generic learning and face synthesis are  complementary. The results also reveal that the performance of FR systems based on the S+V model can further improve with CNN features. 
Future research includes investigating the geometrical structure of the data space in the dictionaries and the corresponding coefficients to improve the discrimination. To reduce reconstruction time, we plan to extend the current S+V model, allowing it to represent larger sparse codes. 

%*********************************
\section*{Reference}
\begin{spacing}{1}
\bibliographystyle{ieee}
\bibliography{refs}
\end{spacing}

\end{document}